\documentclass[11pt]{article}
\usepackage{subcaption}
\usepackage{array}
\usepackage[final]{acl}
\usepackage{booktabs}
\usepackage{multicol}
\usepackage{multirow}
\usepackage{amsmath}
\usepackage{amsfonts}
\usepackage{hyperref}
\usepackage[dvipsnames]{xcolor}
\definecolor{purple}{RGB}{150, 110, 166}
\definecolor{green}{RGB}{130, 179, 102}
\definecolor{orange}{RGB}{215, 155, 0}
\definecolor{yellow}{RGB}{239, 134, 54}
\definecolor{cyan}{RGB}{0, 153, 153}

\usepackage{times}
\usepackage{latexsym}

\usepackage[T1]{fontenc}

\usepackage[utf8]{inputenc}

\usepackage{microtype}

\usepackage{inconsolata}

\usepackage{graphicx}

\title{Proactive for Uncertainty: Cause-Aware Error Diagnosis and Interactive Clarification for Spoken Dialogue Systems}

\author{
  \textbf{Yizhou Peng\thanks{Equal contribution.}\textsuperscript{1}} \quad
  \textbf{Ziyang Ma\footnotemark[1]\textsuperscript{1,2}} \quad
  \textbf{Changsong Liu\textsuperscript{1}} \\
  \textbf{Yi-Wen Chao\textsuperscript{1}} \quad
  \textbf{Xie Chen\textsuperscript{2}} \quad
  \textbf{Eng Siong Chng\textsuperscript{1}} \\[0.5em]
  \textsuperscript{1}Nanyang Technological University, Singapore \quad
  \textsuperscript{2}Shanghai Jiao Tong University, China \\
}

\begin{document}
\maketitle
\begin{abstract}
Cascaded Automatic Speech Recognition -- Large Language Model (ASR-LLM) pipelines remain popular for industrial Spoken Dialogue Systems (SDS), primarily because their decoupled design ensures perceptual verifiability.
However, cascaded systems suffer from error propagation, as transcription failures inevitably cascade to subsequent components, thereby degrading the final interaction quality. Although ASR confidence scores offer a simple filter for unreliable inputs, this approach is fundamentally limited because it typically fails to detect deletion errors or to distinguish between acoustic (inability to hear clearly) and linguistic (inability to understand) mismatches, both of which require targeted recovery strategies.
In this paper, we propose a cause-aware error recovery paradigm that fundamentally rethinks robustness in SDS. Unlike traditional confidence filtering, we introduce a suite of small precision-focused detectors that exploit deep ASR latent representations to disentangle token-level errors into perception, comprehension, and deletion failures. This fine-grained diagnostic intelligence empowers the LLM to orchestrate targeted, multi-turn clarification strategies, effectively transforming ambiguous signals into seamless user interactions.
Experimental results validate the precision of our approach, which more than doubles the recall on domain-shift errors (57.96\% vs. 23.66\%) compared to baselines. Crucially, this diagnostic precision yields up to a 31\% reduction in WER and a 19\% improvement on the downstream task across diverse accents, distortions, and domains.
\end{abstract}

\section{Introduction}

Modern spoken dialogue systems (SDS) generally follow two paradigms: cascaded pipelines or end-to-end (E2E) architectures. While E2E models, such as SALMONN~\cite{SALMONN}, Qwen-Omni~\cite{Qwen-Audio,Qwen2-Audio,Qwen3-Omni}, Step-Audio~\cite{Step-Audio,Step-Audio2}, and full-duplex models~\cite{LSLM,Moshi}, have gained significant traction for their low latency and joint optimization of linguistics and para-linguistics understanding, their intrinsic "black-box" nature poses severe challenges for industrial deployment, where explainability and rapid debugging are paramount.
Consequently, the cascaded pipeline in which user speech is first transcribed by an automatic speech recognition (ASR) model and then processed by a Large Language Model (LLM) remains the mainstream for real-world applications~\cite{funaudiollm,x-talk}.

However, a critical limitation of these pipelines is that the ASR output is not always reliable. In real-world conditions characterized by background noise, accents, or domain-specific vocabulary, transcription errors are inevitable. Without a mechanism to assess whether the transcript matches the user's actual speech, these errors propagate downstream, corrupting intent recognition and response generation~\cite{everson2024_wcn_icl}.
While confidence scoring offers a baseline for error detection, it is insufficient due to model miscalibration~\cite{kuhn2025evaluating} and an inherent inability to capture deletion errors.
Furthermore, if we view the ASR pipeline as a noisy channel, recognition failures stem from two distinct sources that standard confidence metrics conflate. \textit{Perception errors} are \textbf{channel-intrinsic}: the acoustic signal itself is degraded (e.g., background noise, reverberation, or packet loss), so the model cannot \emph{hear} clearly. \textit{Comprehension errors} are \textbf{speech-intrinsic}: the signal is clean, but its content mismatches the recognizer's training distribution, such as rare named entities, specialized jargon, or accented pronunciations, so the model cannot \emph{understand} what it hears. This distinction is critical because the two failure modes demand different recovery strategies: channel-intrinsic failures call for repetition or environmental adjustment, whereas speech-intrinsic failures call for semantic clarification, such as spelling out a word.

We propose a lightweight, cause-aware framework for cascaded spoken dialogue systems that proactively resolves ASR errors. Leveraging diverse embeddings from a frozen ASR model, our model detects token-level errors and diagnoses their underlying causes. These diagnostics guide an LLM in generating targeted clarification strategies to resolve uncertainties through efficient multi-turn dialogue. The resulting clarified transcript prevents error propagation to downstream tasks while minimizing unnecessary user reprompts.
Our contributions are threefold:

\paragraph{Cause-Aware Error Detection.}
We formulate lightweight modules, including \textit{Comprehension}, \textit{Perception}, \textit{Deletion}, and \textit{Distortion Event} detectors, that leverage internal embeddings of a frozen ASR to provide token-level causal diagnostics, crucially including \textit{deletion errors}.

\paragraph{LLM-Driven Clarification.}
We propose a clarification strategy that uses diagnostic error outputs to prompt an LLM with targeted recovery questions. The cause-conditioning strategy enables the system to mimic human-like repair conversations, dynamically adapting its inquiries based on the specific cause of the failures.

\paragraph{Interactive System Evaluation.}
We empirically validate the framework against competing approaches across several domains, demonstrating that targeted clarification significantly reduces WER and directly improves downstream task outcomes across diverse accents and distortions.

\begin{figure*}
    \centering
    \includegraphics[width=1.0\linewidth]{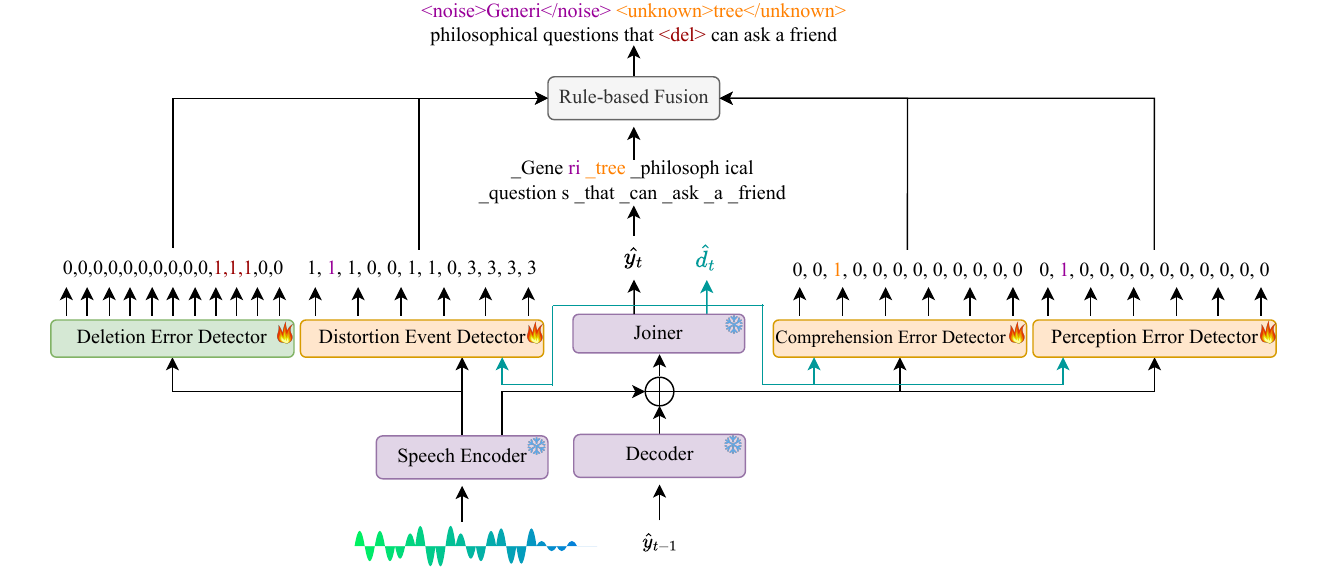}
    \caption{Architecture of the Cause-Aware Error Detector. The Parakeet-tdt ASR backbone (\textcolor{purple}{purple}) feeds four trainable modules: the frame-level \textit{Deletion Error Detector} (\textcolor{green}{green}) and three token-level error detectors (\textcolor{orange}{orange}). The duration prediction of Joiner $\hat{d_t}$ provides the temporal information for token alignment. The deletion error and distortion event detectors use the embeddings output from the speech encoder. In contrast, the comprehension and perception error detectors use joint embeddings that are identical to the input of the joiner network. Outputs are aggregated via rule-based fusion to tag error spans with their specific causes (e.g., \texttt{<noise>}, \texttt{<del>}, and \texttt{<unknown>}), enabling the LLM to generate targeted clarification inquiries before downstream processing.}
    \label{fig:overall-method}
\end{figure*}

\section{Related Work}

\subsection{Sound Event Detection}
Sound Event Detection (SED) evolved from early GMM-HMM approaches to deep architectures like CRNNs, which became a dominant paradigm for temporal modeling~\cite{cakir2017convolutional}. Transformer-based methods (e.g., AST~\cite{gong2021ast}) subsequently advanced global context modeling. Recently, Self-Supervised Learning (SSL) has achieved state-of-the-art performance by leveraging massive unlabeled data~\cite{gong2022ssast}. However, multi-task frameworks that integrate SED directly with ASR remain underexplored, particularly with respect to joint modeling strategies that incorporate environmental awareness without compromising ASR performance.

\subsection{ASR Confidence Estimation and Error Detection}
ASR confidence estimation assesses the reliability of recognition. While early hybrid systems relied on lattice-based posteriors, E2E models necessitate new approaches for subword sequences. Recent research focuses on deriving confidence directly from neural decoder outputs, particularly for RNN-Transducers (RNN-T). Methodologies range from efficient non-parametric entropy measures, such as the Tsallis-entropy confidence of \citet{RNNT-Confidence} (our baseline, Sec. 3.1), to auxiliary neural predictors that frame token-level estimation as a binary classification task~\cite {ogawa2017_error_detection, ogawa2023blstm, wang2021wordlevel}. Furthermore, since downstream tasks typically operate on words, effective aggregation strategies have been validated to propagate these calibrated subword signals to word-level reliability estimates~\cite{oneata2021_slt_confidence, qiu2021learning}.
However, these methods inherently cannot flag deletion errors, as no emitted token exists for frames the model skips.

\subsection{ASR Error Correction and Clarification}
Research addresses ASR noise through downstream adaptation or direct correction. Approaches include adapting SLU models to error patterns~\cite{zhu2018_slu_asr_error_adapt} and leveraging rich structures like word confusion networks~\cite{everson2024_wcn_icl}. Recently, LLM-based post-correction has gained traction, showing promise in contact-center data~\cite{koilakuntla24_interspeech}, multilingual settings~\cite{li24h_interspeech}, and generalized error correction frameworks like HyPoradise~\cite{HyPoradise,ma2024_asr_ec_llm,asano2025_contextual_asr_llm}. In parallel, the dialogue community has established benchmarks (e.g., Qulac, ClarQ) for asking clarifying questions to resolve semantic intent ambiguity~\cite{aliannejadi2019_qulac,kumar_black_2020_clarq,dhole2020_intent_ambiguity}.
However, these interactive methods typically assume reliable text input and lack mechanisms to ground clarification decisions in ASR-derived evidence.
Notably, the challenge of error recovery in spoken dialogue predates the LLM era: foundational work by \citet{bohus2007error} and \citet{skantze2007error} established principled frameworks for error awareness and recovery strategies in conversational interfaces, demonstrating that effective recovery requires balancing interaction efficiency against the risk of misunderstanding. Our work revisits this challenge in the context of modern cascaded ASR-LLM pipelines, grounding clarification decisions in fine-grained, cause-aware diagnostics rather than coarse confidence signals.

\section{Method}

\subsection[Background: Entropy-based Error Detection]{Entropy-based Error Detection}
\label{sec:entropy_baseline}
To mitigate the overconfidence often observed in RNN-T models using standard softmax or Shannon entropy, we adopt the training-free confidence measure proposed by~\citet{RNNT-Confidence} based on \textit{Tsallis entropy}. This approach introduces a tunable entropic index $\alpha \in (0, 1)$ to better manage sensitivity of the probability distribution. The Tsallis Confidence for a token $u$ over a vocabulary of size $|\mathcal{V}|$ is defined as:

\begin{equation}
    \label{eq:tsallis_exp_conf}
    C_{\text{token}}(u) = \frac{\exp\left( \frac{|\mathcal{V}|^{1-\alpha} - \sum_{\mathcal{V}} p_v^\alpha}{1-\alpha} \right) - 1}{\exp\left( \frac{|\mathcal{V}|^{1-\alpha} - 1}{1-\alpha} \right) - 1}.
\end{equation}

\noindent For a word composed of token sequences $\{u_1, \dots, u_M\}$, the word-level confidence is the arithmetic mean of its constituent token scores:
\begin{equation}
    \label{eq:word_aggregation}
    C_{\text{word}}(W) = \frac{1}{M} \sum_{k=1}^{M} C_{\text{token}}(u_k).
\end{equation}

\noindent Finally, a word is classified as an \textit{Error} if its confidence falls below a calibrated threshold $\tau$. The threshold $\tau$ is tuned on each set to satisfy a specific strictness constraint, maximizing error detection recall while ensuring the False Positive Rate (FPR) remains below a target limit $\beta$.

\subsection{Token and Duration Transducer}
We utilize the \textbf{Parakeet-tdt-0.6b-v2} (en)\footnote{\url{https://huggingface.co/nvidia/parakeet-tdt-0.6b-v2}} model, which pairs a FastConformer encoder (80ms/frame) with a Token-and-Duration Transducer (TDT) decoder~\cite{xu2023tdt}. TDT optimizes inference by jointly predicting the subword token $\hat{y}_t \in \mathcal{V}$ (where $|\mathcal{V}| = 1024$) and its duration $\hat{d}_t \in \{0, 1, 2, 3, 4\}$.

During greedy decoding, the model selects the optimal pair $(\hat{y_t}, \hat{d_t})$ via the Joiner and advances the encoder state by $\hat{d}_t$ frames ($t \leftarrow t + \hat{d}_t$), allowing it to skip redundant frames for faster inference.

\begin{figure*}
    \centering
    \includegraphics[width=0.90\linewidth]{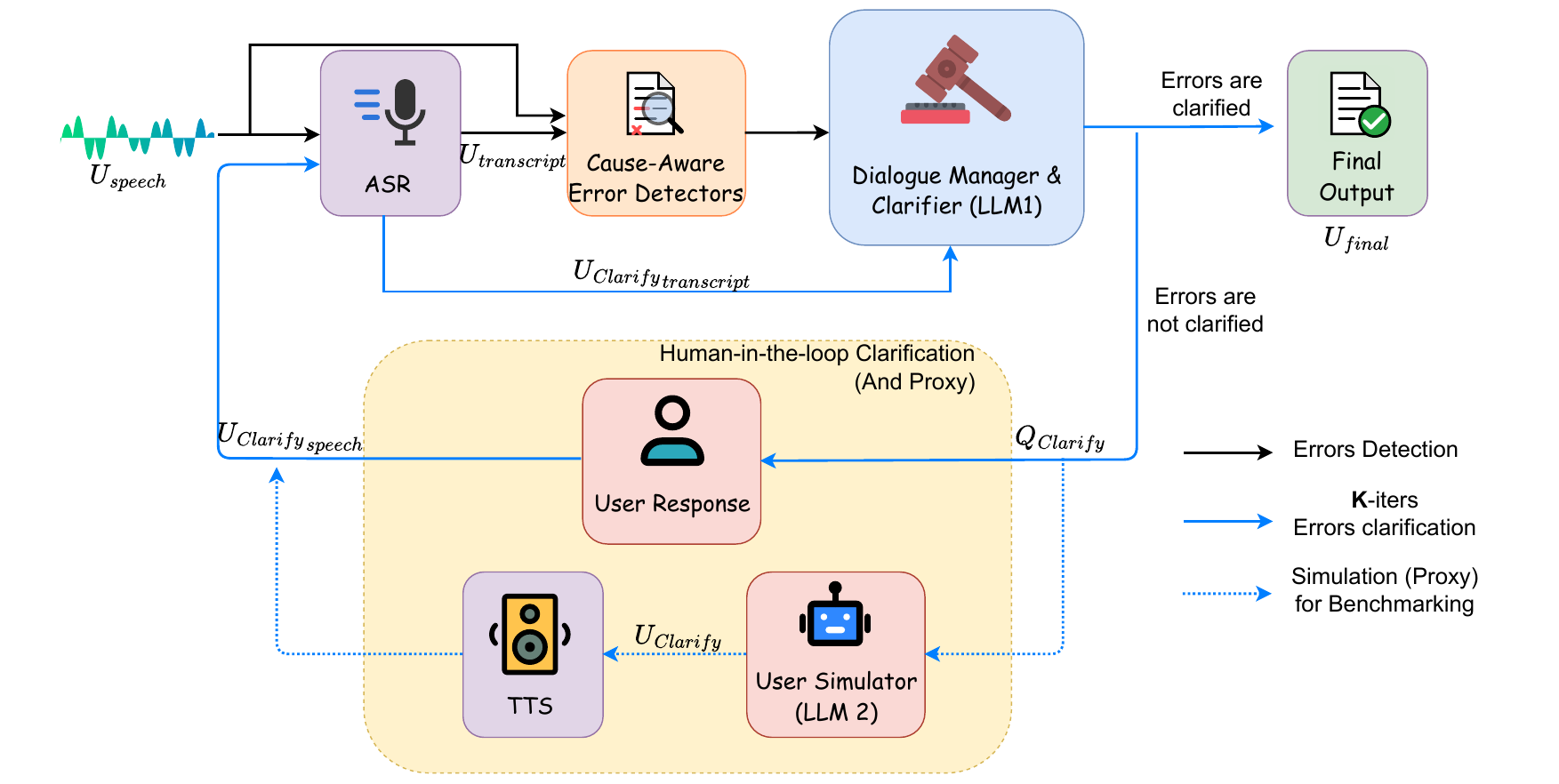}
    \caption{The complete Cause-Aware Error Recovery pipeline. The process begins with the User's initial inquiry, which is transcribed by the ASR and immediately analyzed by Cause-Aware Error Detectors. If failures are identified, the Dialogue Manager (LLM1) orchestrates targeted error-correction strategies to resolve ambiguities, prompting the user for clarification directly via speech input (solid arrows). The dashed pathway indicates the automated benchmarking setup, where a User Simulator (LLM2) and TTS are employed solely as a Proxy to simulate this interaction for experimental evaluation.}
    \label{fig:clarify_bench_tts}
    \vspace{-0.5em}
\end{figure*}

\subsection{ASR Error Detectors}
To directly flag recognition failures, we deploy three specialized detectors: the \textbf{Deletion}, \textbf{Comprehension}, and \textbf{Perception} Error Detectors, as shown in Figure~\ref{fig:overall-method}. The Perception detector targets channel-intrinsic failures, while the Comprehension detector targets speech-intrinsic ones. The Comprehension and Perception detectors operate at the token level and are designed to capture \textit{substitution} and \textit{insertion} errors. The Deletion detector addresses the complementary case of \textit{missing} content. For word-level evaluation, a word is labeled as an error if any of its constituent tokens are classified as erroneous.

To maintain efficiency, all detectors share a unified classification formulation. Given an input embedding $\mathbf{x}$ and a task-specific label set $\mathcal{C}$, each detector predicts the class $\hat{c}$ via:
\begin{equation}
    \label{eq:general_classifier}
    \hat{c} = \operatorname*{argmax}_{c \in \mathcal{C}} P(c \mid \mathbf{x})
\end{equation}

\noindent \textbf{Comprehension \& Perception Detectors:} These modules operate at the \textbf{token level} and utilize the TDT joint embedding $\mathbf{z}_{u}^{\text{joint}}$ as input ($\mathbf{x} = \mathbf{z}_{u}^{\text{joint}}$). This embedding is defined as:
\begin{equation}
    \label{eq:joint}
    \mathbf{z}_{u}^{\text{joint}} = \mathbf{W}_{e} \mathbf{h}_{t_u}^{\text{enc}} + \mathbf{W}_{d} \mathbf{h}_{u}^{\text{dec}} + \mathbf{b}_{z}
\end{equation}
where $\mathbf{h}_{t_u}^{\text{enc}}$ and $\mathbf{h}_{u}^{\text{dec}}$ are the encoder and decoder states, respectively.
Both use the binary class set $\mathcal{C} = \{\text{Correct}, \text{Error}\}$.

\noindent \textbf{Deletion Detector:} Unlike the token-level modules, this detector operates at the \textbf{frame level} to identify speech content missed by the decoder. It utilizes the raw encoder outputs as input ($\mathbf{x} = \mathbf{h}_{t}^{\text{enc}}$) with the class set $\mathcal{C} = \{\text{Correct}, \text{Deletion}\}$. To distinguish valid deletions from silence, we apply a masking condition based on TDT emissions:
\begin{equation}
    e_t = \mathbb{I}(\hat{y}_t = \text{Del}) \cdot \mathbb{I}(\text{No Emission at } t)
\end{equation}
\noindent To count distinct deletion events, we aggregate consecutive positive flags into single instances.

\subsection{Distortion Event Detector}
Distinct from detecting ASR errors, the \textit{Distortion Event Detector} characterizes the acoustic environment at the \textbf{token level} (see Appendix~\ref{ap:disentangle} for the empirical rationale behind this separation).
It utilizes the same formulation (Eq.~\ref{eq:general_classifier}) but operates on the Encoder embeddings ($\mathbf{x} = \mathbf{h}_{t_u}^{\text{enc}}$) aligned to the token duration.

We define a set of six environmental states: $\mathcal{C}_{event}$ (detailed in Appendix~\ref{ap:events}), comprising: \textbf{Clean, Interference, Noise, Room Impulse Response (RIR), Packet Loss, Missing}. This ensures the distortion classification is strictly grounded in the specific acoustic frame aligned with the token emission.

\subsection{Cause-Aware Clarification Pipeline}

We propose an iterative framework with up to $K$ rounds that integrates a frozen ASR backbone, error detectors, and an LLM-based Dialogue Manager into a unified recovery loop, as shown in Figure~\ref{fig:clarify_bench_tts}.

Given a user utterance $U_{speech}$, the ASR model generates a transcript $U_{transcript}$ while the detector suite produces a structured error profile $E = \{ Y_{comp}, Y_{perc}, Y_{del}, Y_{event} \}$. This profile identifies both the location and root cause of failures. The Dialogue Manager uses $(U_{transcript}, E)$ to select a targeted strategy: prompting repetition or seeking a quieter room if Perception errors are flagged; or asking clarifying questions or requesting that the word be spelled if comprehension errors are detected.
When multiple detectors flag the same token, we apply a deterministic priority rule ($\text{Comprehension} > \text{Perception} > \text{Deletion}$): linguistic mismatches take precedence because acoustic-oriented strategies (e.g., ``repeat'') would not resolve a speech-intrinsic, domain-mismatch error.

For clarification, the system engages in a $K$-iteration dialogue to resolve errors. In each step $k$, the \textit{Dialogue Manager \& Clarifier} formulates a query $Q_{clarify}^{(k)}$ addressing the specific unresolved errors in $E^{(k)}$. The user's spoken response is re-transcribed and re-analyzed to refine the hypothesis. This cycle repeats until the detectors confirm a clean transcript or the maximum $K$ is reached, yielding $U_{final}$. Note that for dataset-level benchmarking, we simulate the "User" role using an LLM to generate clarifying text and a TTS model to synthesize the speech response, as shown in the dashed arrow in Figure~\ref{fig:clarify_bench_tts}.

\section{Experimental Setup}

\subsection{Datasets and Data Construction}
We utilize three distinct corpora to cover a wide range of linguistic and acoustic conditions for training and evaluation: \textbf{LibriSpeech}~\cite{panayotov2015librispeech} as a standard reading benchmark, \textbf{SPGISpeech2}~\cite{grossman2025spgispeech} for financial domain teleconference speech, and \textbf{AESRC2020}~\cite{shi2021accented} for diverse English accents. To illustrate the robustness of the proposed method, we further include Out-of-Domain (OOD) evaluation sets: \textbf{Gigaspeech}~\cite{gigaspeech}, \textbf{WSJ}~\cite{wsj}, \textbf{Openhermes}~\cite{openhermes}, and \textbf{Alpaca}~\cite{alpaca}, where the final two subsets~\cite{Audiobench} are also used for spoken dialogue evaluation. (Data details in Appendix~\ref{app:datasets}).

\paragraph{Comprehension Task.}
To capture intrinsic model failures caused by domain shifts or accents, we employ the raw, clean recordings from these corpora paired with their groundtruth transcripts.
\begin{itemize}
    \item \textbf{Labeling Strategy:} We align the clean ASR hypothesis $H_{clean}$ against the groundtruth $Y_{GT}$. A token is labeled as a \textit{Comprehension Error} if it belongs to a word aligned as a substitution or insertion.
\end{itemize}

\paragraph{Perception Task and Event Detection.}
To target errors induced by acoustic distortion, we generate LibriSpeech-distortions and SPGI2-distortions by applying \textbf{nine} synthetic acoustic conditions to the clean subsets, namely Interference, Missing, Multi-dist (No RIR), Multi-dist (RIR), Noise, Noise (Partial), Packet Loss, RIR, RIR+Noise.
\begin{itemize}
    \item \textbf{Labeling Strategy:} We employ a differential labeling approach, treating the clean hypothesis $H_{clean}$ as a pseudo-oracle. By aligning the distorted hypothesis $H_{dist}$ against it, we label a token as a \textit{Perception Error} if it is different in $H_{dist}$, and as a \textit{Deletion Error} if it is present in $H_{clean}$ but missing in $H_{dist}$.
\end{itemize}

\begin{table*}[t]
    \centering
    \small
    \begin{tabular}{lccccc}
        \toprule
        & \multicolumn{2}{c}{\textbf{Deletion Error Detector}} & & \multicolumn{2}{c}{\textbf{Percep. / Compr. Error Detector}} \\
        \cmidrule{2-3} \cmidrule{5-6}
        \textbf{Test Condition} & \textbf{FPR ($\downarrow$)} & \textbf{Recall ($\uparrow$)} & & \textbf{FPR ($\downarrow$)} & \textbf{Recall ($\uparrow$)} \\
        \midrule
        \multicolumn{6}{l}{\textit{\textbf{Perception Task}}} \\
        ~~Interference & 0.62 & 62.44 & & 4.79 & 66.29 \\
        ~~Missing & 1.87 & 92.42 & & 5.52 & 64.10 \\
        ~~Multi-dist (No RIR) & 1.27 & 82.23 & & 5.72 & 65.82 \\
        ~~Multi-dist (RIR) & 1.46 & 77.52 & & 8.50 & 72.75 \\
        ~~Noise & 0.33 & 46.17 & & 5.63 & 73.14 \\
        ~~Noise (Partial) & 0.28 & 38.84 & & 4.75 & 66.84 \\
        ~~Packet Loss & 0.32 & 39.70 & & 4.37 & 62.14 \\
        ~~RIR & 0.29 & 23.60 & & 4.61 & 62.57 \\
        ~~RIR + Noise & 0.47 & 44.36 & & 8.37 & 74.10 \\
        \textbf{Avg. } & \textbf{0.77} & \textbf{56.36} & & \textbf{5.81} & \textbf{67.53} \\
        \midrule
        \multicolumn{6}{l}{\textit{\textbf{Comprehension Task}}} \\
        ~~AESRC-Test & \multicolumn{2}{c}{-} & & 1.61 & 47.98 \\
        ~~SPGI2-Test & \multicolumn{2}{c}{-} & & 1.78 & 70.26 \\
        \bottomrule
    \end{tabular}
    \caption{Token-level Error Detection Evaluation. False Positive Rates (\textbf{FPR}) and Recall Rates are reported. Perception Task evaluates the synthetic distorted SPGI2-Test using both the \textbf{Deletion} and \textbf{Percep}tion Error Detector. Comprehension Task evaluates two clean sets using the \textbf{Compr}ehension Error Detector. }
    \label{tab:token_level_results}
    \vspace{-0.5em}
\end{table*}

\subsection{Model Configuration}
\label{sec:hyper_params}

We freeze the \textbf{Parakeet-tdt-0.6b-v2} backbone. All four detectors employ an identical architecture: a \textbf{5-layer 1D-CNN}, selected based on systematic ablation over linear, MLP, CNN, and Transformer architectures (Appendix~\ref{ap:ablation_architecture}). The \textit{Distortion Event Detector} uses a 6-class output head (defined in $\mathcal{C}_{event}$), while the other modules use binary classification heads. Each detector comprises $\approx$10M parameters. Full training hyperparameters are provided in Appendix~\ref{ap:training_details}.

For K-round clarification and Spoken Dialogue Evaluation, we set $K=3$, and utilize \textit{GPT-5.2} to power the Dialogue Manager (LLM1), the User Simulator (LLM2), and the Model-as-a-Judge (MaJ) scorer, alongside CosyVoice3~\cite{cosyvoice3} for TTS, in a \textbf{fixed-reference-voice setup}: all simulated user replies are synthesized with the same clean reference speaker, irrespective of the original speaker's voice (an accent-preserving alternative is evaluated in Sec.~\ref{sec:sim_validity}). We implement a rigorous \textbf{quarantine protocol} to prevent context leakage: LLM2 is provided with the user's specific target intent, whereas LLM1 is explicitly denied access to this groundtruth. This ensures that the Dialogue Manager must resolve ambiguities dynamically during conversation, rather than cheating by relying on hidden context.
Every system-level experiment is executed as three independent full-pipeline runs, and results in Sec.~\ref{sec:werr_eval}--\ref{sec:sim_validity} are reported as empirical mean $\pm$ 95\% confidence intervals ($t$-distribution, $n{=}3$).

\section{Quantitative Evaluations}

We evaluate the proposed framework in two phases: component-level detection accuracy and system-level interactive effectiveness evaluation.

We evaluate the framework in two phases: component-level detection accuracy (\textit{Recall}, \textit{FPR}, \textit{F1}) and system-level recovery effectiveness (\textit{WERR}, \textit{MaJ} scoring). Metric definitions are in Appendix~\ref{app:metrics}.

\begin{table*}[t]
    \centering
    \small
    \begin{tabular}{lcc|cc|cc}
        \toprule
        & \multicolumn{2}{c|}{\textbf{Deletion Error Detector}} &  \multicolumn{4}{c}{\textbf{Percep. / Compr. Error Detector}} \\
        & & & \multicolumn{2}{c|}{\textbf{Entropy Baseline}} & \multicolumn{2}{c}{\textbf{Proposed Method}} \\
        \cmidrule{2-7}
        \textbf{Test Condition} & \textbf{FPR} ($\downarrow$) & \textbf{Recall} ($\uparrow$) & \textbf{FPR}\textsuperscript{\dag} ($\downarrow$) & \textbf{Recall} ($\uparrow$) & \textbf{FPR} ($\downarrow$) & \textbf{Recall} ($\uparrow$) \\
        \midrule
        \multicolumn{7}{l}{\textit{\textbf{Perception Task }(SPGI2-Test-Distorted)} } \\
        ~~Interference & 4.82 & 38.67 & 8.32 & 37.99 & 8.31 & \textbf{47.48} \\
        ~~Missing & 6.00 & 63.37 & 9.78 & 37.01 & 9.78 & \textbf{49.05} \\
        ~~Multi-dist (No RIR) & 5.27 & 56.81 & 9.82 & 39.41 & 9.82 & \textbf{50.06} \\
        ~~Multi-dist (RIR) & 6.69 & 47.42 & 14.28 & 53.33 & 14.25 & \textbf{61.56} \\
        ~~Noise & 3.85 & 22.35 & 9.39 & 41.40 & 9.39 & \textbf{54.37} \\
        ~~Noise (Partial) & 3.67 & 21.94 & 8.04 & 37.28 & 8.04 & \textbf{47.93} \\
        ~~Packet Loss & 3.34 & 24.41 & 7.21 & 33.57 & 7.21 & \textbf{43.88} \\
        ~~RIR & 3.31 & 20.79 & 8.31 & 37.84 & 8.31 & \textbf{48.24} \\
        ~~RIR + Noise & 3.99 & 20.21 & 12.78 & 51.03 & 12.78 & \textbf{60.01} \\
        \textit{\textbf{Avg. }} & \textit{4.70} & \textit{44.67} & \textit{9.66} & \textit{41.57} & \textit{9.64} & \textbf{\textit{52.26}} \\
        \midrule
        \multicolumn{7}{l}{\textit{\textbf{Comprehension Task} (Clean)}} \\
        ~~AESRC-Test & \multicolumn{2}{c|}{-} & 1.13 & 13.26 & 1.13 & \textbf{39.38} \\
        ~~SPGI2-Test & \multicolumn{2}{c|}{-} & 4.01 & 23.66 & 3.98 & \textbf{57.96} \\

        \midrule
        \multicolumn{7}{l}{\textit{\textbf{Out-of-Domain}}} \\
        ~~Gigaspeech & 3.35 & 23.15 & 5.37 & 26.02 & 5.37 & \textbf{29.70} \\
        ~~OpenHermes  & 0.20 & 0.00$^\ddagger$ & 2.00 & \textbf{75.00} & 2.07 & 43.63 \\
        ~~Alpaca & 0.30 & 0.00$^\ddagger$ & 4.35 & 53.33 & 4.65 & \textbf{71.05} \\
        ~~wsj-eval92 & 0.00 & 2.52 & 0.69 & 7.37 & 0.69 & \textbf{29.72} \\
        \bottomrule
        \multicolumn{7}{l}{\footnotesize{$^\dag$ The threshold was tuned to align the Baseline FPR with the Proposed Method for fair comparison.}} \\
        \multicolumn{7}{l}{\footnotesize{$^\ddagger$ 0\% recall on OpenHermes/Alpaca due to negligible deletion errors ($<$2 instances).}}
    \end{tabular}
    \caption{Word-level Error Detection Evaluation. False Positive Rates (\textbf{FPR}) and Recall Rates are reported. We benchmark the proposed \textbf{Deletion}, \textbf{Perception}, and \textbf{Comprehension} Error Detectors against the \textbf{Entropy-based Baseline}. Evaluations span three conditions: \textit{Perception}, \textit{Comprehension}, and \textit{Out-of-Domain (OOD)}. Note that the Entropy method serves as the comparator for the \textbf{Perception} or \textbf{Comprehension} Error Detector on the corresponding task, and for the combination of the two detectors under OOD conditions.}
    \label{tab:word_level_benchmark}

\end{table*}

\subsection{Token Level Evaluation}
We perform token-level error detection on the in-domain test sets, including AESRC-Test and SPGI2-Test, because we have only the artificial token-level label under these conditions.
Table~\ref{tab:token_level_results} presents the performance of the Deletion, Perception, and Comprehension error detectors. The results demonstrate the efficacy of decoupling error detection tasks to achieve targeted sensitivity.

\paragraph{Perception Task Performance}
In acoustic distortion tasks, the Perception Error Detector is robust (67.53\% average recall, 5.81\% FPR) and most effective in noisy environments (> 73\% Recall). Conversely, the Deletion Error Detector prioritizes precision (0.77\% FPR). While it excels at identifying missing segments (92.42\% Recall), performance drops significantly in RIR conditions (23.60\% Recall), as reverberation smears the sharp signal boundaries required to identify deletions.

\paragraph{Comprehension Task Performance}
The Comprehension Error Detector maintains a conservative profile on linguistic tasks ($<$1.8\% FPR). Recall is domain-dependent: the detector performs well on the finance-specific SPGI2-Test (70.26\%). Still, it shows moderate recall on the accented AESRC (47.98\%), indicating the model is more sensitive to terminology errors than accent deviations.

\subsection{Word Level Evaluation}
To evaluate the system's practical utility, we aggregate token-level predictions into word-level error flags. Table~\ref{tab:word_level_benchmark} benchmarks our proposed detectors against the \textit{Tsallis Entropy Baseline} across three diverse testing conditions. To ensure a fair comparison, we tune the baseline's confidence threshold to match the FPR of the proposed method.

\paragraph{Comparison with Entropy Baseline}
Our proposed method consistently outperforms the entropy baseline across both tasks. In the \textit{Perception Task}, it achieves a superior average Recall of \textbf{52.26\%} compared with 41.57\%, indicating that our learned modules capture signal artifacts more effectively than simple model confidence. This performance gap becomes even more profound in the \textit{Comprehension Task}: our method drastically improves Recall from 23.66\% to \textbf{57.96\%} on \textit{SPGI2-Test} and from 13.26\% to \textbf{39.38\%} on \textit{AESRC-Test}. These results demonstrate that, although the entropy baseline fails when the ASR model is "confidently wrong" under a \textbf{clean recording environment}, our supervised detector successfully identifies these intrinsic failures.

\paragraph{Deletion Detector Specialization}
The \textbf{Deletion Error Detector} continues to exhibit a precision-oriented behavior at the word level. Its average FPR on the Perception task is significantly lower (\textbf{4.70\%}) than the Perception Error Detector (\textbf{9.64\%}). Despite its narrower scope, it achieves a respectable \textbf{Average Recall of 44.67\%}, with particular effectiveness in the \textit{Missing} condition (63.37\%), validating its role as a specialized trigger for handling missing words.

\begin{table}[h]
    \centering
    \small
    \begin{tabular}{lcc}
        \toprule
        \textbf{Distortion Type} & \textbf{F1 Score ($\uparrow$)} & \textbf{Acc ($\uparrow$)} \\
        \midrule
        Clean & 91.89 & 88.92 \\
        Noisy & 15.79 & 35.85 \\
        RIR & 6.76 & 9.59 \\
        Interference & 29.26 & 38.79 \\
        Packet Loss & 26.32 & 25.44 \\
        Missing & 78.58 & 73.67 \\
        \midrule
        \textbf{Macro F1 / Micro Acc.} & \textbf{41.43} & 85.49 \\
        \bottomrule
        \\
    \end{tabular}
    \caption{Token-level Evaluation Results for Sound Event Detection. F1 score (scale of 100) and \textbf{Acc}uracy(\%) are reported for each distortion type. Macro-F1 is reported for Average.}
    \label{tab:sed_evaluation}
\end{table}

\paragraph{OOD Generalization}
We further evaluate robustness on datasets unseen during training (Gigaspeech, OpenHermes, Alpaca, WSJ). Our proposed method demonstrates strong generalization, notably achieving \textbf{71.05\%} Recall on Alpaca (vs. 53.33\% baseline) and \textbf{29.72\%} on wsj-eval92 (vs. 7.37\% baseline). In contrast, the Deletion Detector shows near-zero recall on OpenHermes and Alpaca; as noted in Table~\ref{tab:word_level_benchmark}, this is an expected artifact of the test sets containing almost no deletion errors, rather than a model failure.

\begin{table}[t]
    \centering
    \small
    \setlength{\tabcolsep}{3pt}
    \begin{tabular}{l c ccc}
        \toprule
        \multirow{2}{*}{\textbf{Setup}} & \textbf{Init.} & \multicolumn{3}{c}{\textbf{Refinement (WER $\downarrow$)}} \\
        \cmidrule(lr){3-5}
         & \textbf{ASR}$^{\S}$ & \textbf{Step 1} & \textbf{Step 2} & \textbf{Step 3} \\
        \midrule
        \multicolumn{5}{l}{\textit{\textbf{WSJ-eval92}}} \\
        ~~HyPoradise~\shortcite{HyPoradise}$^\dag$ & 7.60 & 7.30 & -- & 6.30 \\
        ~~\textbf{Proposed} & \textbf{4.23} & \textbf{4.02} & 3.90 & {\textbf{3.77{\scriptsize$\pm$0.17}}} \\
        \midrule
        \multicolumn{5}{l}{\textit{\textbf{Gigaspeech} {(single run$^{\ddagger}$)}}} \\
        ~~\textbf{Proposed} & 14.57 & 12.88 & 12.47 & \textbf{12.28} \\
        \midrule
        \multicolumn{5}{l}{\textit{\textbf{SPGI-noise}}} \\
        ~~\textbf{Proposed} & 17.57 & 15.24 & 13.85 & {\textbf{12.12{\scriptsize$\pm$0.22}}} \\
        \midrule
        \multicolumn{5}{l}{\textit{\textbf{AESRC-Indian}}} \\
        ~~\textbf{Proposed} & 6.08 & 4.91 & 4.79 & {\textbf{4.72{\scriptsize$\pm$0.05}}} \\
        {~~~~w/ accented TTS$^{\P}$} & {6.08} & {--} & {--} & {5.15{\scriptsize$\pm$0.07}} \\
        \bottomrule
        \multicolumn{5}{l}{\footnotesize{$^\dag$\textit{HyPoradise}: Step 1 is 0-shot; Step 3 is 10-shot;}}\\
        \multicolumn{5}{l}{\footnotesize{~~cross-backbone reference only.}}\\
        \multicolumn{5}{l}{\footnotesize{{$^{\S}$Greedy one-best decoding.}}}\\
        \multicolumn{5}{l}{\footnotesize{{$^{\ddagger}$Single run; excluded from N=3 re-runs.}}}\\
        \multicolumn{5}{l}{\footnotesize{{$^{\P}$Replies cloned from the original accented speaker}}}\\
        \multicolumn{5}{l}{\footnotesize{{~~(Sec.~\ref{sec:sim_validity}).}}}\\
    \end{tabular}
    \caption{WER comparison for 3 rounds of interactive clarification. {HyPoradise~\cite{HyPoradise} is retained as a cross-backbone reference on WSJ-eval92{; the same-backbone, non-interactive N-best+LLM baseline is reported in Appendix Table~\ref{tab:ap_baseline}}. Step 1--2 values are from run 1; final-round (Step 3) values for WSJ, SPGI-noise, and AESRC are mean $\pm$ 95\% CI over N=3 independent full-pipeline runs.}}
    \label{tab:wer_comparison}
    \raggedright

    \end{table}

\subsection{Speech Event Detection}
Table~\ref{tab:sed_evaluation} evaluates the multi-class classification capability of the \textit{Distortion Event Detector}. Macro-F1, the appropriate summary for a multi-class attribution task, is \textbf{41.43}, with strongly bimodal per-class behavior: it reliably identifies \textbf{Clean} (F1 91.89) and \textbf{Missing} (F1 78.58), but performs poorly on fine-grained distortion types. The overall accuracy of 85.49\%, reported for completeness, is dominated by the majority \textit{Clean} class and should not be read as a measure of attribution quality. The high performance on the \textit{Missing} class further corroborates the reliability of our deletion-detection strategy by distinguishing explicit signal loss from other artifacts.

\begin{figure}[t]
    \centering
    \includegraphics[width=1.0\linewidth]{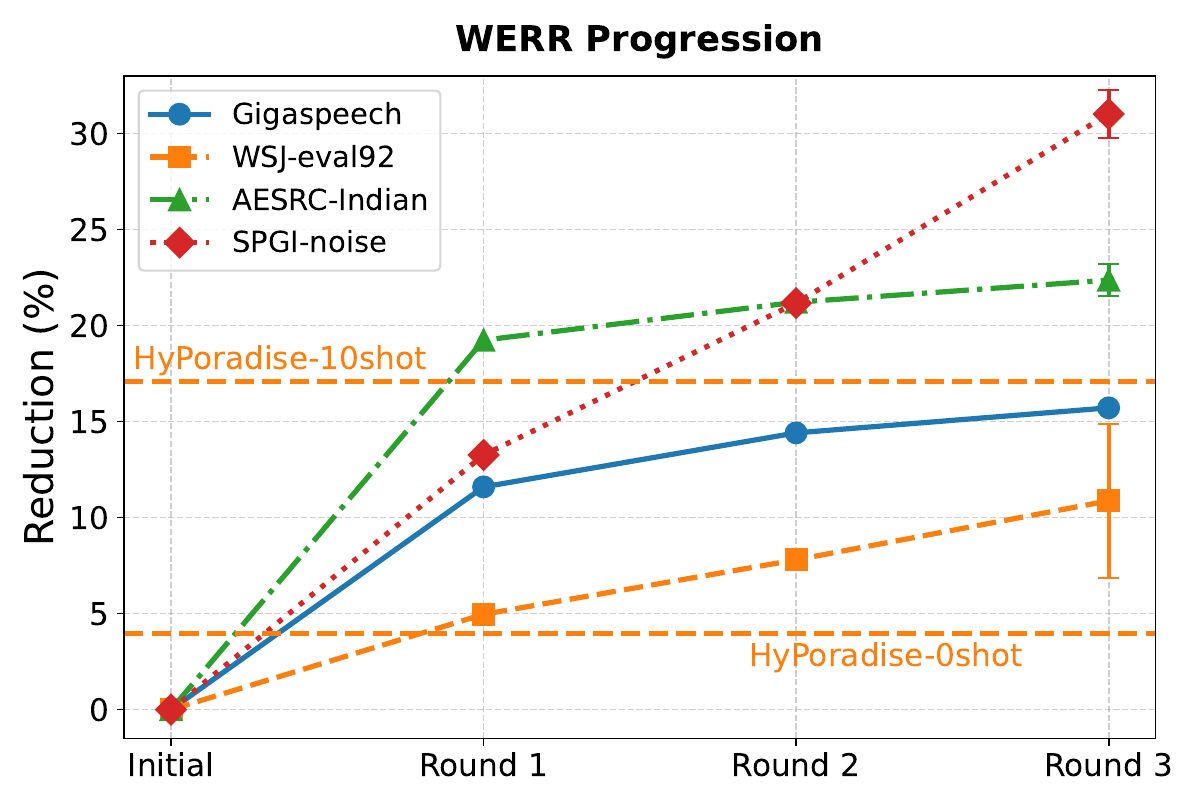}
    \caption{WERR performance of 3-round Clarification for the four test sets. Two reference horizontal \textcolor{yellow}{dashed lines} are results extracted from HyPoradise~\citep{HyPoradise} for WSJ-eval92. {Final-round points use the N=3 mean WERs of Table~\ref{tab:wer_comparison} (error bars: 95\% CIs propagated to WERR); Gigaspeech is a single run.}}
    \label{fig:k-round-werr}
\end{figure}

However, the model struggles to differentiate between fine-grained additive distortions. Performance drops significantly for \textbf{RIR} (F1=6.76) and \textbf{Noisy} (F1=15.79) conditions. This suggests that while the ASR backbone's embeddings strongly encode the \textit{presence} or \textit{absence} of speech (Clean vs. Missing), the representations for different environmental corruptions (e.g., reverb vs. background noise) are highly entangled, making specific cause classification challenging at the token level.
A detailed misdiagnosis analysis on the SPGI2-Test-Noise subset confirms that 69.4\% of noise-induced errors are correctly attributed, with the main failure mode being conservative ``Clean'' classification rather than cross-category confusion (Appendix~\ref{ap:failure_analysis}).
{Crucially, imperfect attribution does not break the downstream recovery: conditioning per-utterance $\Delta$WER on attribution correctness (Appendix~\ref{ap:attribution}) shows that misattributed utterances still retain roughly 60\% of the correctly-attributed gain, and the dominant residual cost comes instead from error-bearing utterances that receive no tag at all.}

\subsection[3-Round Clarification]{3-Round {Clarification}}
\label{sec:werr_eval}
We validate the complete recovery pipeline through {three independent full-pipeline runs} of 3-round clarification on specific hard-to-recognize subsets: the \textit{AESRC-Indian accent}, the \textit{SPGI-noise} condition, and a filtered version of Gigaspeech ($\approx$11k samples) where utterances requiring complex numeric normalization were excluded{; final-round WERs are reported as mean $\pm$ 95\% CI ($t$-distribution, $n{=}3$), except Gigaspeech, which is evaluated in a single run due to its size and API cost}.
Table~\ref{tab:wer_comparison} reports the absolute WER, while Figure~\ref{fig:k-round-werr} illustrates the relative WERR progression.

\paragraph{Robustness on in/out domain Subsets.}
The system achieves consistent gains across all subsets, with WERR peaking at {31.0\%} on \textit{SPGI} and {22.4\%} on \textit{AESRC}, demonstrating robust recovery from both severe environmental noise and firm accent shifts. The WERR increases monotonically with each round, confirming that our multi-turn strategy effectively corrects residual errors.

{
\paragraph{Same-Backbone Baseline.}
We further compare against a non-interactive, HyPoradise-style post-correction baseline on the same backbone (beam-4 N-best + a single \textit{gpt-5.2} correction call; Appendix~\ref{ap:nbest_baseline}). On clean, read speech (WSJ), the single-shot baseline is statistically indistinguishable from our interactive system, whereas under accent and distortion the interactive system leads by large, significant margins ($p<10^{-4}$ in every run). Interaction is thus not a universally better correction method, but the right tool precisely for the conditions this paper targets, where the acoustic signal rather than the language model is the bottleneck.
}

\paragraph{Baseline Comparison.}
On \textit{WSJ-eval92}, our WERR surpasses HyPoradise's~\shortcite{HyPoradise} zero-shot baseline but trails their 10-shot result. However, they optimize a much weaker baseline (WER=7.60\%), making gains easier. Our method refines a stronger backbone (WER=4.23\%) to a superior result (WER={\textbf{3.77\%}}), significantly outperforming their best absolute result (WER=6.30\%).
\begin{table}[t]
    \centering
    \small
    \setlength{\tabcolsep}{4pt}
    \begin{tabular}{l l c c}
        \toprule
        \multicolumn{2}{c}{\textbf{Experimental Setting}} & \multicolumn{2}{c}{\textbf{Dial Quality (MaJ)}} \\
        \cmidrule(lr){1-2} \cmidrule(lr){3-4}
        \textbf{Input Source} & \textbf{WER} & \textbf{OpenH} & \textbf{Alpaca} \\
        \midrule
        \multicolumn{4}{l}{\textit{Reference Upper Bounds}} \\
        ~~GT & Oracle & {86.73{\scriptsize$\pm$1.74}} & {84.47{\scriptsize$\pm$2.74}} \\
        ~~ASR & Clean & {86.00{\scriptsize$\pm$3.10}} & {83.87{\scriptsize$\pm$0.29}} \\
        \midrule
        \multicolumn{4}{l}{\textit{Distortion \& Recovery}} \\
        ~~ASR & Dist$\dag$ & {75.27{\scriptsize$\pm$1.88}} & {69.07{\scriptsize$\pm$2.29}} \\
        ~~\textbf{Clarified} & 3-round & {\textbf{82.73{\scriptsize$\pm$2.74}}} & {\textbf{82.53{\scriptsize$\pm$3.73}}} \\
        \bottomrule
    \multicolumn{4}{l}{\footnotesize{$\dag$ The same setting as SPGI2-Test-Multi-dist (No RIR)}}
    \end{tabular}
    \caption{Downstream Dialogue Evaluation on \textbf{OpenH}ermes and \textbf{Alpaca} test sets. MaJ scores (scaled 0-100) compare our proposed \textbf{3-Round Clarification} against the \textbf{ASR-Dist} baseline, with \textbf{ASR-Clean} and \textbf{G}round\textbf{T}ruth inputs serving as upper bounds. {All values are mean $\pm$ 95\% CI over N=3 independent full-pipeline runs, scored by the GPT judge.}}
    \label{tab:dialogue_settings}
    \vspace{-1.0em}
\end{table}

\subsection{Spoken Dialogue Evaluation}
\label{sec:sd_eval}
We assess downstream utility using the MaJ score on instruction-following tasks {over three independent runs}, as shown in Table~\ref{tab:dialogue_settings}. While distortions significantly degrade performance, our cause-aware error detection and clarification pipeline restores the score to {\textbf{82.73} and \textbf{82.53}}. This recovery effectively bridges the gap to the clean baseline ({$\sim$86, 84}){: clarification recovers $78\%\pm15\%$ of the Dist$\rightarrow$GT judge-score gap under the GPT judge ($85\%\pm14\%$ under the independent judge of Sec.~\ref{sec:sim_validity})}, demonstrating that our semantic corrections directly translate to improved task execution.
A supplementary qualitative evaluation of 141 clarification rounds confirms that approximately 60\% of system queries are judged both natural and effective by MaJ scoring (Appendix~\ref{ap:naturalness}).

{
\subsection{Validity of the Simulated Interaction}
\label{sec:sim_validity}
Our benchmark replaces the human interlocutor with three proxies: an LLM judge, an LLM user simulator, and a TTS voice. A natural concern is that each proxy makes the problem easier than deployment with real users. To address this concern, we quantify the optimism introduced by each proxy along its own dimension (Appendix~\ref{ap:sim_validity_details}).

{
\paragraph{Scoring.}
We re-scored every frozen output with an independent judge family (\textit{Gemini-3.5-Flash}), and the conclusions are unchanged: the two judges agree closely, GPT is the \emph{harsher} of the two on every condition, and the independent judge attributes an equal or larger share of the recovery to clarification. The reported gains are therefore not an artifact of GPT judging GPT.

\paragraph{User behavior.}
We replaced the fully cooperative simulator with mildly and strongly uncooperative variants (intent-only knowledge, imprecise paraphrasing, and limited patience), and downstream MaJ remains statistically unchanged: the gains do not depend on oracle cooperation.

\paragraph{Speech channel.}
We additionally synthesized the clarification replies in the original speaker's accented voice rather than the clean fixed voice; the AESRC recovery shrinks by ${\approx}0.5$ WER (Table~\ref{tab:wer_comparison}). The benefit remains significant in every run.

Together, these controls bound the simulator's optimism rather than assume it away: the judge is not self-preferring, the user need not be an oracle, and the idealized speech channel accounts for roughly a third of the accent-condition gain.
}
}

\section{Conclusion}
\label{sec:Conclusion}
In this work, we presented a framework to make Spoken Dialogue Systems \textbf{proactive against uncertainty}. We found that traditional confidence-based filtering is insufficient for cascaded systems, as it fails to distinguish between acoustic-perception failures and linguistic-comprehension gaps. To address this, we introduced a \textbf{cause-aware diagnosis mechanism} that leverages specialized detectors to pinpoint the root cause of errors, doubling error detection recall with comparable precision under domain shifts compared with baselines. By coupling this diagnostic precision with \textbf{interactive clarification}, our approach enables LLMs to resolve ambiguities through user-in-the-loop engagement, offering \textbf{context-sensitive mitigation strategies tailored to the specific failure reason}, ranging from requests for repetition to suggestions for environmental adjustments.
Experimental results confirm the efficacy of this paradigm, demonstrating a {\textbf{31\% reduction in WER}} and a {\textbf{19\% improvement in downstream utility}}. Ultimately, this validates that moving from passive error filtration to proactive, cause-aware resolution is essential for building resilient dialogue agents.

\section*{Limitations}
\label{sec:limits}
This study has several limitations that also imply practical risks. We deliberately constrained the LLM to edit only detector-flagged tokens to isolate detector precision, which may cause the system to miss unflagged errors and may lead us to underestimate LLM–detector synergy. All experiments target English speech only; other languages may introduce challenges such as tonal distinctions or agglutinative morphology.
Training data is drawn from SPGISpeech2 (financial) and AESRC (Accented English) with simulated distortions, so performance may degrade under real-world conditions (e.g., non-stationary noise, overlapping speakers) or unseen domains. We report automatic metrics without subjective human validation, leaving open the question of whether cause-aware clarifications feel natural to users.

\section*{Acknowledgments}
This research is supported by the RIE2025 Industry Alignment Fund - Industry Collaboration Projects (IAF-ICP) (Award I2301E0026), administered by A*STAR, as well as supported by Alibaba Group and NTU Singapore through Alibaba-NTU Global e-Sustainability CorpLab (ANGEL).

\section*{Ethics Statement}
Our experiments use only publicly available datasets, including community benchmarks and competition datasets that are claimed to have been desensitized, and we do not collect new data or conduct human evaluation.

\paragraph{Potential risks.}
The primary risks of this work relate to (i) misinterpretation or misuse of model outputs, and (ii) performance disparities across domains, accents, or conditions represented in the benchmarks. We clearly report limitations and potential problems in \hyperref[sec:limits]{Limitations}.

\paragraph{AI assistance disclosure.}
We utilized AI tools, including ChatGPT, Gemini, Claude, and Grammarly, to assist with language editing, initial code generation, and README preparation. In accordance with ACL policy, we disclose the scope of this use; the authors have manually verified all outputs and retain full responsibility for the final work.
\bibliography{custom}

\appendix

\section{Appendix}
\label{sec:appendix}

\subsection{Open Source and Reproducibility}
\label{ap:open_source}
To facilitate future research and ensure the reproducibility of our results, {we release the complete codebase, covering the whole pipeline: data preprocessing, distortion simulation, model training, inference scripts, and the interactive LLM clarification module.\footnote{{\url{https://github.com/pengyizhou/Cause-Aware-Error-Detection-and-Correction}}} The model weights of the classifiers are also included in the repository.}

\subsection{Dataset Details}
\label{app:datasets}
\begin{table*}[t]
    \centering
    \footnotesize
    \begin{tabular}{lllc >{\centering\arraybackslash}p{2.6cm}}
        \toprule
        \textbf{Exp. Task} & \textbf{Dataset} & \textbf{Split} & \textbf{\# Samples} & \textbf{Duration (hrs)} \\
        \midrule
        \multirow{9}{*}{\textbf{Comprehension}} & \multirow{3}{*}{AESRC2020} & Train & 119k & 142.0 \\
         & & Valid & 5.6k & 6.7 \\
         & & Test & 14.5k & 16.8 \\
         \cmidrule{2-5}
         & \multirow{2}{*}{LibriSpeech} & Train (clean-100) & 28.5k & 100.6 \\
         & & Dev (clean+other) & 5.5k & 10.5 \\
         \cmidrule{2-5}
         & \multirow{3}{*}{SPGI2-clean} & Train & 159k & 421.0 \\
         & & Valid & 6.9k & 18.3 \\
         & & Test & 7.6k & 20.5 \\
        \midrule
        \midrule
        \multirow{6}{*}{\textbf{Perception}} & \multirow{2}{*}{LibriSpeech-distortions} & Train & 4.8k$^\dagger$ & 16.7$^\dagger$ \\
         & & Dev & 0.9k$^\dagger$ & 1.7$^\dagger$ \\
         \cmidrule{2-5}
         & \multirow{3}{*}{SPGI2-distortions} & Train & 26.5k$^\dagger$ & 70.0$^\dagger$ \\
         & & Valid & 1.1k$^\dagger$ & 3.0$^\dagger$ \\
         & & Test & 1.3k$^\dagger$ & 3.5$^\dagger$ \\
        \midrule
        \midrule
        \multirow{4}{*}{\textbf{Out-of-Domain}} & Gigaspeech & Test & 19.9k & 20 \\
         & WSJ-eval92 & Test & 333 & 0.7 \\
         & OpenHermes & Test & 100 & 0.17 \\
         & Alpaca & Test & 100 & 0.12 \\
        \bottomrule
        \multicolumn{5}{l}{\footnotesize{$^\dagger$ Averaged per nine distortion subset.}
        }
    \end{tabular}%
    \caption{Statistics of the datasets used for Comprehension (Linguistic) and Perception (Acoustic) experiments. For the Perception datasets, we generate 9 types of distortions: Interference, Noise, Noise-partial, RIR, RIR-Noise, Packet Loss, Missing, Multi-dist (w/o RIR), and Multi-dist (w/ RIR). Note that the statistics for distorted sets are reported \textit{per distortion type}. For Out-of-Domain evaluation, we only have test sets, while their training data is not included in our experiments.}
    \label{tab:datasets}
\end{table*}
Table~\ref{tab:datasets} summarizes the statistics for all datasets used in our experiments. Below we detail the source corpora and distortion generation protocols.

\subsubsection{Comprehension Corpora}
We utilize three distinct datasets to represent varying levels of linguistic difficulty:
\begin{itemize}
    \item \textbf{AESRC2020}: Sourced from the \textit{Accented English Speech Recognition Challenge}, this dataset contains studio-quality recordings covering eight distinct English accents: American, British, Chinese, Indian, Japanese, Korean, Portuguese, and Russian.
    \item \textbf{LibriSpeech}: A standard benchmark derived from audiobooks. We train on \texttt{train-clean-100} and validate on the union of \texttt{dev-clean} and \texttt{dev-other}, representing high-quality, standard reading speech.
    \item \textbf{SPGI2-clean}: A subset of the SPGISpeech2 corpus, consisting of multi-speaker spontaneous conversational teleconference recordings from the financial domain. We selected segments shorter than 20 seconds, yielding 460 hours from the original 3780-hour corpus.
\end{itemize}

\subsubsection{Out-of-Domain (OOD) Corpora} To evaluate the robustness of our error recovery pipeline against unseen domains and acoustic conditions, we incorporate four distinct Out-of-Domain (OOD) datasets. These sets introduce varying levels of difficulty, ranging from standard reading tasks to complex, instruction-following dialogue scenarios.

\begin{itemize} \item \textbf{Gigaspeech}: A large-scale multi-domain corpus derived from audiobooks, podcasts, and YouTube. For our OOD evaluation, we use the test split with very long segments ($>$30\,s) removed, yielding 19.9k samples (20 hours); this is the set used for the baseline WER of Table~\ref{tab:ASR-results} and the OOD detection results of Table~\ref{tab:word_level_benchmark}. This dataset challenges the model with diverse recording environments and spontaneous speaking styles.

\item \textbf{WSJ-eval92}: The standard evaluation set from the Wall Street Journal (CSR-I) corpus. It consists of high-quality read speech of news text. We use this as a benchmark for "clean" but domain-shifted performance, enabling comparison with established baselines such as HyPoradise.

\item \textbf{OpenHermes}: A massive collection of synthetic instruction-following datasets, initially designed for fine-tuning Large Language Models (LLMs). We use the 100-sample subset from AudioBench~\cite{Audiobench}.

\item \textbf{Alpaca}: Similar to OpenHermes, this dataset consists of 52,000 instructions generated by OpenAI's text-davinci-003. We use the 100-sample subset from AudioBench~\cite{Audiobench}.

\end{itemize}

\subsection{Perception Corpora Construction}
\label{ap:events}
For the Perception task, we introduced five fundamental distortion types: \textbf{Noise, Interference, Reverberation (RIR), Packet Loss,} and \textbf{Signal Missing}.
We define these distortions as follows:

\paragraph{Noise (Additive).}
This represents environmental background sounds that degrade the signal-to-noise ratio (SNR) of the target speech. We sample non-speech noise clips from the \textbf{MUSAN} corpus~\cite{snyder2015musan} and add them to the clean audio at SNRs uniformly sampled from the range $[-5, 20]$ dB.

\paragraph{Reverberation (RIR).}
Reverberation simulates the multipath propagation of sound in enclosed spaces. We convolve the clean speech with Room Impulse Responses (RIRs) selected from the \textbf{MUSAN} corpus~\cite{snyder2015musan}, specifically targeting "medium" and "large" room profiles to introduce significant late reflections and spectral coloration.

\paragraph{Interference (Competing Speaker).}
This event simulates the "cocktail party problem" where a secondary speaker talks simultaneously. We select interfering speech segments (distinct from the target speaker) and mix them additively with the target audio. The mixing SNR is sampled from $[5, 20]$ dB to ensure the target remains dominant but challenged.

\paragraph{Packet Loss (Low-Bitrate Compression).}
To simulate transmission artifacts and information loss typical of unstable network conditions (e.g., VoIP), we employ aggressive lossy compression. We utilize \texttt{FFmpeg} with the \textbf{Opus} codec~\cite{valin2012opus}, a standard for real-time communication. Randomly selected segments are compressed to extremely low bitrates $b \in \{1, 2, 4, 8\}$ kbps using the following command:
\begin{center}
\texttt{ffmpeg -i \$in -c:a libopus -b:a \$b \$out}
\end{center}
This process introduces blocking artifacts and spectral truncation analogous to severe packet loss concealment.

\paragraph{Missing (Temporal Masking).}
This models hardware failures or mute errors where audio data is completely lost. We simulate this by strictly zeroing out (masking) a randomly selected contiguous segment of the audio waveform (setting amplitude to 0).

By applying these to the clean subsets of LibriSpeech and SPGI2, we generated \textbf{nine} distinct test conditions representing single-source, partial, and composite distortions:
\begin{enumerate}
    \item \textbf{Single-Source:} Noise (entire segment), RIR (entire segment).
    \item \textbf{Partial/Composite:} Interference, Packet Loss, Missing, Noise-partial (applied to random sub-segments), RIR+Noise, Multi-distortion (w/ RIR), and Multi-distortion (w/o RIR).
\end{enumerate}

\paragraph{Data Sampling.}
To construct the training and evaluation sets for these nine conditions, we maintained the original Train/Valid/Test splits from the source corpora. For each specific distortion subset, we randomly sampled \textbf{1/6} of the corresponding clean source samples. This sampling ratio ensures sufficient coverage while intentionally creating data overlap across different distortion types, preventing the model from overfitting to a specific subset size.

\subsection{Training Details}
\label{ap:training_details}

Both the encoder outputs (projected via the internal $1024\!\to\!640$ projection) and the joint embeddings share a unified dimension of $d_{model} = 640$. For the \textit{Tsallis entropy} baseline, we set the entropic index $\alpha=0.33$.

All four detectors use a kernel size of 5 and a dropout rate of 0.2. Models are trained with the AdamW optimizer for 40 epochs, using a learning rate of $2 \times 10^{-4}$, weight decay of $0.01$, and a batch size of 192. All models were trained on a single NVIDIA A40-48GB GPU. The final model was derived by averaging the top 5 checkpoints (saved at 500-step intervals) sorted by validation set error recall.

For the LLM-based evaluation pipeline, we use GPT-5.2 (snapshot: 2025-12-11).

\subsection{Evaluation Metrics}
\label{app:metrics}

To comprehensively assess the performance of the proposed framework, we employ a diverse set of metrics covering error detection, event classification, and dialogue recovery.

\subsubsection{Error Detection Metrics}
For the binary error detectors (Comprehension, Perception, Deletion) at both the token and word levels, we prioritize the trade-off between sensitivity and false alarms.
We define the \textbf{Positive} class as an \textit{Error} and the \textbf{Negative} class as a \textit{Correct} token. Consequently, the confusion matrix components are defined as:
True Positive (\textbf{TP}): An actual error correctly detected by the model.
False Negative (\textbf{FN}): An actual error missed by the model.
False Positive (\textbf{FP}): A correct token incorrectly flagged as an error.
True Negative (\textbf{TN}): A correct token classified adequately as correct.

Consequently, Recall and False Positive Rate (FPR) are defined as:
\begin{itemize}
    \item \textbf{Recall (Sensitivity):} Measures the proportion of actual errors correctly identified by the detector.
    \begin{equation}
        \text{Recall} = \frac{\text{TP}}{\text{TP} + \text{FN}}
    \end{equation}
    \item \textbf{FPR (Precisions):} Measures the proportion of correct tokens that are incorrectly flagged as errors. Low FPR is critical to prevent unnecessary user interruptions.
    \begin{equation}
        \text{FPR} = \frac{\text{FP}}{\text{FP} + \text{TN}}
    \end{equation}
\end{itemize}

\subsubsection{Events Classification Metrics}
For the multi-class \textit{Distortion Event Detector}, we utilize standard classification metrics evaluated at the token level:
\begin{itemize}
    \item \textbf{Accuracy:} The ratio of correctly classified frames to the total number of frames.
    \item \textbf{F1 Score:} The mean of precision and recall, averaged across all six environmental classes to account for class imbalance.
\end{itemize}

\subsubsection{Error Correction Metrics}
To evaluate the improvement in transcript quality after the clarification loop, we report the following metrics:
\begin{itemize}
    \item \textbf{Word Error Rate (WER):} The standard metric for ASR performance, calculated as $\text{WER} = \frac{S + D + I}{N}$, where $S, D, I$ are substitutions, deletions, and insertions, and $N$ is the number of words in the reference.
    \item \textbf{WER Reduction (WERR):} Quantifies the relative improvement achieved by the system compared to the initial frozen ASR baseline.
\end{itemize}

\subsubsection{Dialogue Quality Metrics}
To assess the generative performance of the clarification pipeline, we employ a \textbf{Model-as-a-Judge (MaJ)} framework using OpenAI's GPT-5.2. Adopting the evaluation protocol from AudioBench~\cite{Audiobench}\footnote{\url{https://github.com/AudioLLMs/AudioBench}}, the judge assigns a quality score (scale 0--100) based on the semantic alignment of the \textit{Predicted Answer} with the reference \textit{Groundtruth Question} and \textit{Groundtruth Answer} triplet.

\subsection{Ablation on Classifier Architecture}
\label{ap:ablation_architecture}

A key design choice in our framework is the architecture of the lightweight classification heads attached to the frozen ASR backbone. To justify the selection of the 1D-CNN classifier described in Section~\ref{sec:hyper_params}, we conducted an ablation study over nine architectural configurations, evaluating each on the Perception Error Detection task using the SPGI2-distortion validation set.

\begin{table}[t]
    \centering
    \footnotesize
    \setlength{\tabcolsep}{3.5pt}
    \begin{tabular}{lccc}
        \toprule
        \textbf{Architecture} & Clean F1 & Noisy F1 & Macro F1 \\
        \midrule
        \multicolumn{4}{l}{\textit{Non-contextual}} \\
        ~~Linear (640$\to$2) & 0.968 & 0.112 & 0.540 \\
        ~~2-layer MLP & 0.969 & 0.234 & 0.602 \\
        ~~MLP + enc-proj & 0.970 & 0.337 & 0.654 \\
        \midrule
        \multicolumn{4}{l}{\textit{CNN variants}} \\
        ~~CNN ($k\!=\!5$, $L\!=\!2$) & 0.975 & 0.542 & 0.759 \\
        ~~CNN ($k\!=\!5$, $L\!=\!4$) & \textbf{0.976} & \textbf{0.555} & \textbf{0.766} \\
        ~~CNN ($k\!=\!5$, $L\!=\!5$) & 0.975 & 0.542 & 0.759 \\
        \midrule
        \multicolumn{4}{l}{\textit{Hybrid / Transformer}} \\
        ~~MLP + enc-proj + CNN & 0.971 & 0.382 & 0.677 \\
        ~~\quad+ full-range & 0.971 & 0.369 & 0.670 \\
        ~~Conv-TF & 0.971 & 0.404 & 0.688 \\
        \bottomrule
    \end{tabular}
    \caption{Architecture ablation for the Perception Error Detector on the SPGI2-distortion validation set. ``enc-proj'' denotes the encoder's internal 1024$\to$640 projection. Conv-TF uses 2 CNN layers followed by 3 Transformer layers with 8 attention heads. All models use a hidden dimension of 640 and are trained for 40 epochs. The ``full-range'' setting means that we label the whole perception error (full range of frames) with the label, while other settings only label the first frame.}
    \label{tab:ablation_architecture}
\end{table}

\paragraph{Results.}
Table~\ref{tab:ablation_architecture} reports the findings. Three clear trends emerge:

\textit{(1) Local context is essential.} The jump from non-contextual baselines (Linear/MLP, Macro F1 $\leq$ 0.654) to the CNN variants (Macro F1 $\geq$ 0.759) demonstrates that temporal context from neighboring frames is critical for distinguishing real errors from transient fluctuations. A single-frame classifier achieves high accuracy on the dominant Clean class but catastrophically fails on the minority Noisy class (F1 = 0.112).

\textit{(2) CNN outperforms Transformer on this task.} The Conv-Transformer hybrid (Macro F1 = 0.688) underperforms the pure CNN (0.766) despite having more parameters and global attention. We attribute this to the short-range, localized nature of acoustic distortion patterns: error signals are typically confined to a few adjacent frames, matching the CNN's inductive bias for local pattern detection. The Transformer's global attention introduces unnecessary capacity that leads to overfitting on the relatively small training set.

\textit{(3) Depth saturates at 4 layers.} Among the CNN variants, $L\!=\!4$ achieves the best Macro F1 (0.766), while $L\!=\!5$ shows no further gain (0.759). This suggests that a receptive field spanning approximately 17 frames ($\sim$1.4\,s at 80\,ms/frame) is sufficient to capture the relevant distortion context without overfitting. We adopt the $L\!=\!5$ configuration in the final system for a marginal receptive field increase as a safety margin, though the $L\!=\!4$ variant is equally viable.

\subsection{Rationale for separating Perception and Distortion Detectors}
\label{ap:disentangle}
Our architecture explicitly disentangles the \textit{Perception Error Detector} from the \textit{Distortion Event Detector} due to their distinct feature dependencies. Through ablation studies, we observed conflicting requirements for optimal performance:

\begin{itemize}
    \item \textbf{Perception Error Detection} requires understanding the alignment between the acoustic signal and the linguistic history to identify misrecognized content. Consequently, it achieves superior performance when utilizing the \textbf{Joint embedding} defined as Eq.\ref{eq:joint}, which fuses the encoder's acoustic representations with the decoder's label history.
    \item \textbf{Distortion Event Detection}, conversely, relies on preserving raw signal characteristics to identify environmental noise, artifacts, or blurring. Our experiments show that the Encoder embedding retains this necessary acoustic information, whereas the Joint embedding tends to abstract it during fusion.
\end{itemize}

Empirically, attempting to unify these tasks into a single model using Joint embeddings resulted in a significant performance degradation for distortion detection.

\begin{table}[t]
    \centering
    \small
    \setlength{\tabcolsep}{4pt}
    \begin{tabular}{l cccccc}
        \toprule
        & \multicolumn{6}{c}{\textbf{Predicted Distortion}} \\
        \cmidrule(lr){2-7}
        \textbf{True} & Noise & RIR & Interf. & PktLoss & Missing & Clean \\
        \midrule
        Noise & \textbf{758} & 47 & 11 & 6 & 3 & 268 \\
        (\%) & \textbf{69.4} & 4.3 & 1.0 & 0.5 & 0.3 & 24.5 \\
        \bottomrule
    \end{tabular}
    \caption{Cause attribution analysis on SPGI-Test-Noise. For 1,093 true noise-induced perception errors, we report how the Distortion Event Detector classified the underlying cause. The dominant correct attribution (\textbf{69.4\%} Noise) validates the detector's reliability under adverse conditions.}
    \label{tab:misdiagnosis}
\end{table}

\subsection{Failure Analysis and Misdiagnosis}
\label{ap:failure_analysis}

To assess the robustness of the cause-aware detectors in adversarial conditions, we analyze the \textit{misdiagnosis rate}: cases where the detector correctly identifies an error but attributes it to the wrong environmental cause. This analysis is critical because an incorrect cause attribution can lead to an inappropriate recovery strategy (e.g., asking the user to repeat when the issue is network packet loss).

\paragraph{Setup.}
We focus on the SPGI-Test-Noise subset, which contains 1,093 true perception errors caused by additive noise. For each correctly detected error token, we compare the \textit{Distortion Event Detector}'s predicted cause against the known ground-truth distortion type.

\paragraph{Results.}
Table~\ref{tab:misdiagnosis} presents the cause attribution breakdown. Of the 1,093 noise-induced errors, \textbf{69.4\%} are correctly attributed to the ``Noise'' category, confirming that the detector reliably identifies the dominant environmental cause. The most common misattribution is to ``Clean'' (24.5\%), which reflects the conservative detection strategy discussed in Section~\ref{ap:event_detection}: at the token emission instant, the acoustic features occasionally retain sufficient clarity for the model to perceive them as undistorted, even though the surrounding context is noisy. Critically, misattributions to other distortion types (RIR: 4.3\%, Interference: 1.0\%, Packet Loss: 0.5\%, Missing: 0.3\%) are rare, indicating that the detector does not confuse fundamentally different failure modes.

\begin{table}[htp]
    \centering
    \footnotesize
    \begin{tabular}{lc}
        \toprule
        \textbf{Condition / Accent} & \textbf{WER (\%)} \\
        \midrule
        \multicolumn{2}{l}{\textit{\textbf{Perception Task} (SPGI-Test)}} \\
        Clean Baseline & 13.61 \\
        \midrule
        Interference & 14.92 \\
        Missing & 31.42 \\
        Packet Loss & 12.23 \\
        Noise & 17.57 \\
        Noise (Partial) & 14.62 \\
        RIR & 13.77 \\
        RIR + Noise & 21.34 \\
        Multi-dist (No RIR) & 21.21 \\
        Multi-dist (RIR) & 26.25 \\
        \midrule
        \midrule
        \multicolumn{2}{l}{\textit{\textbf{Comprehension Task} (AESRC-Test)}} \\
        \textbf{Average} & \textbf{5.36} \\
        \midrule
        American & 3.74 \\
        British & 0.72 \\
        Chinese & 5.11 \\
        Indian & 6.08 \\
        Japanese & 9.74 \\
        Korean & 6.83 \\
        Portuguese & 5.08 \\
        Russian & 7.81 \\
        \midrule
        {\textit{\textbf{Out-of-Domain Test Sets}}} \\
        Gigaspeech & 11.69 \\
        OpenHermes$\dag$ & 4.00 \\
        Alpaca$\dag$ & 4.12 \\
        wsj-eval92 & 4.23 \\
        \bottomrule
        \multicolumn{2}{l}{\footnotesize{$\dag$Synthesized speech}}
    \end{tabular}%
    \caption{WER results of Parakeet-tdt-v2 model on the in-domain and out-of-domain test sets. The AESRC-Test \textbf{Average} is the corpus-level (micro) WER pooled over all eight accents, not the mean of the per-accent rows.}
    \label{tab:ASR-results}
\end{table}

\paragraph{Impact on Recovery Strategy.}
The 24.5\% ``Clean'' misattribution does \textit{not} result in a missed error: the token is still flagged by the Perception Error Detector. Instead, the system falls back to a generic ``please repeat'' strategy rather than the more targeted ``move to a quieter environment'' suggestion. While suboptimal, this fallback remains a valid recovery action. In contrast, cross-category confusion (e.g., Noise $\to$ Interference) is sufficiently rare ($<$5\%) that inappropriate strategies (e.g., ``wait for the other speaker to finish'') are unlikely to confuse users in practice.

\subsection{ASR Results over Test Sets}
Table~\ref{tab:ASR-results} details the baseline performance (WER) of the Parakeet-tdt-v2 backbone across all evaluation conditions.

In the \textbf{Perception Task}, although the model remains robust to mild interference, severe signal degradations, such as \textit{Signal Missing} (31.42\%) and \textit{Multi-distortion} (26.25\%), cause substantial performance drops relative to the clean baseline (13.61\%). In the \textbf{Comprehension Task}, we observe significant variance across accents: while standard British and American English yield low WERs ($<$4\%), heavily accented speech such as \textit{Japanese} (9.74\%) and \textit{Russian} (7.81\%) presents a much more complex challenge. Finally, the \textbf{OOD} results indicate that, although the model excels on clean reading and instructional speech (WSJ, OpenHermes, and Alpaca), it still struggles with the spontaneous and diverse nature of Gigaspeech (11.69\%).
\begin{figure*}[t]
    \centering
    \begin{subfigure}[b]{0.48\linewidth}
        \centering
        \includegraphics[width=1.0\textwidth]{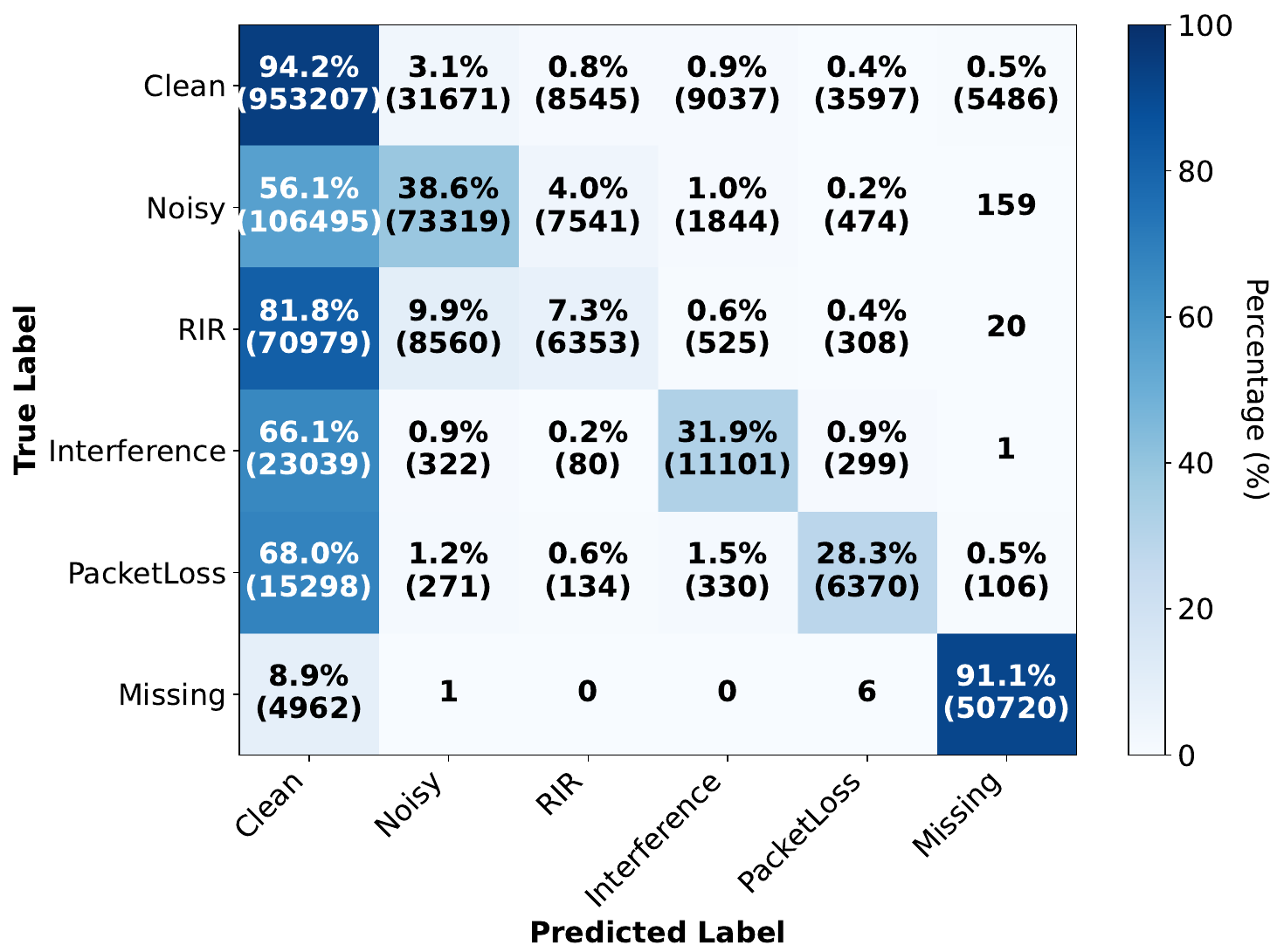}
        \caption{Frame-level confusion matrix}
        \label{fig:confusion_frame}
    \end{subfigure}
    \hfill
    \begin{subfigure}[b]{0.48\linewidth}
        \centering
        \includegraphics[width=1.0\textwidth]{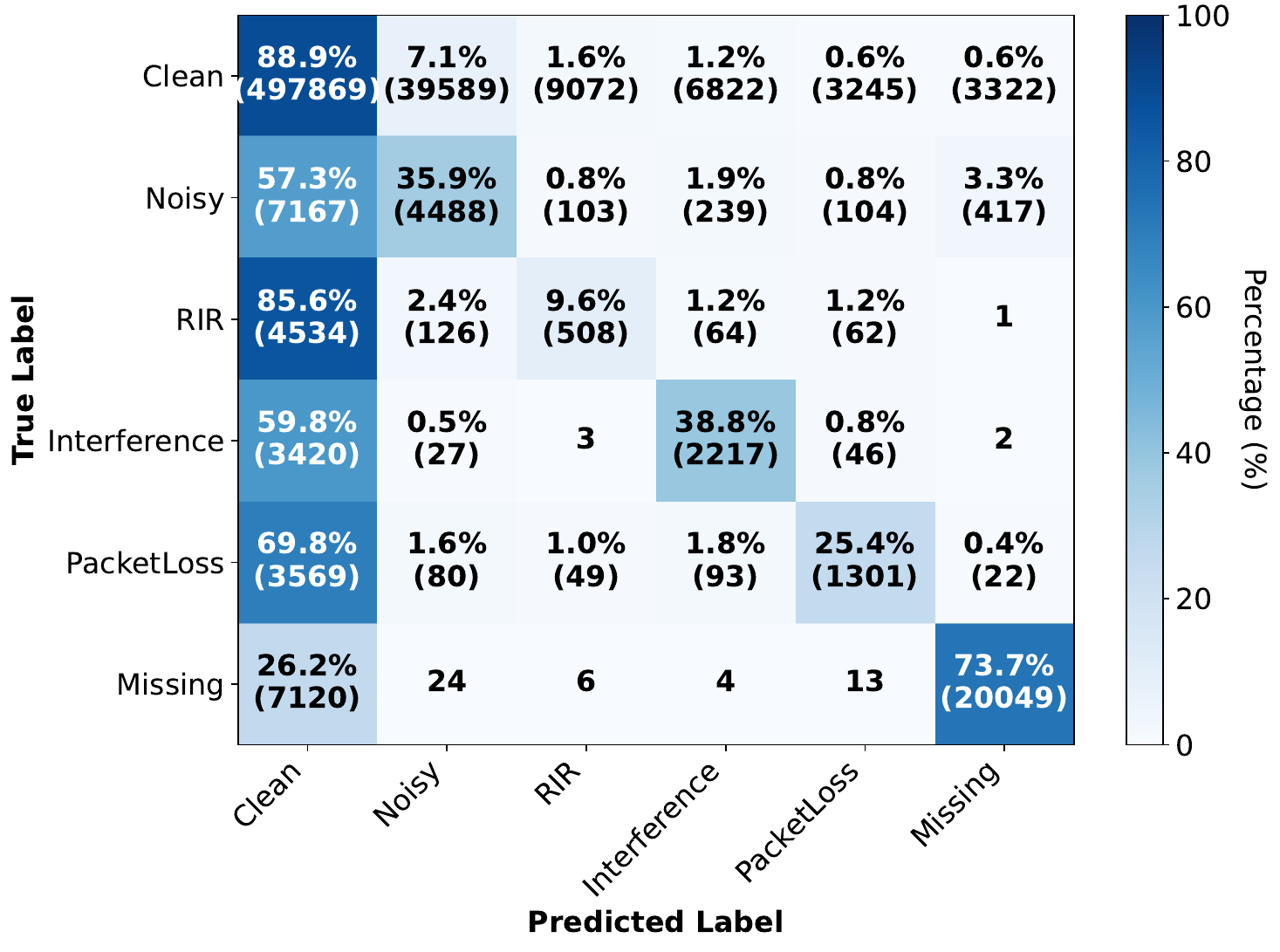}
        \caption{Token-level confusion matrix}
        \label{fig:confusion_token}
    \end{subfigure}
    \caption{Confusion matrices showing classification performance for distortion events. (a) Frame-level predictions aggregated across all events. (b) Token-level predictions aggregated across all events. Both matrices are normalized by row to yield percentages, facilitating comparison of performance across classes with different sample sizes.}
    \label{fig:confusion_matrices}
\end{figure*}
\begin{figure*}[t]
    \centering
    \includegraphics[width=1.0\linewidth]{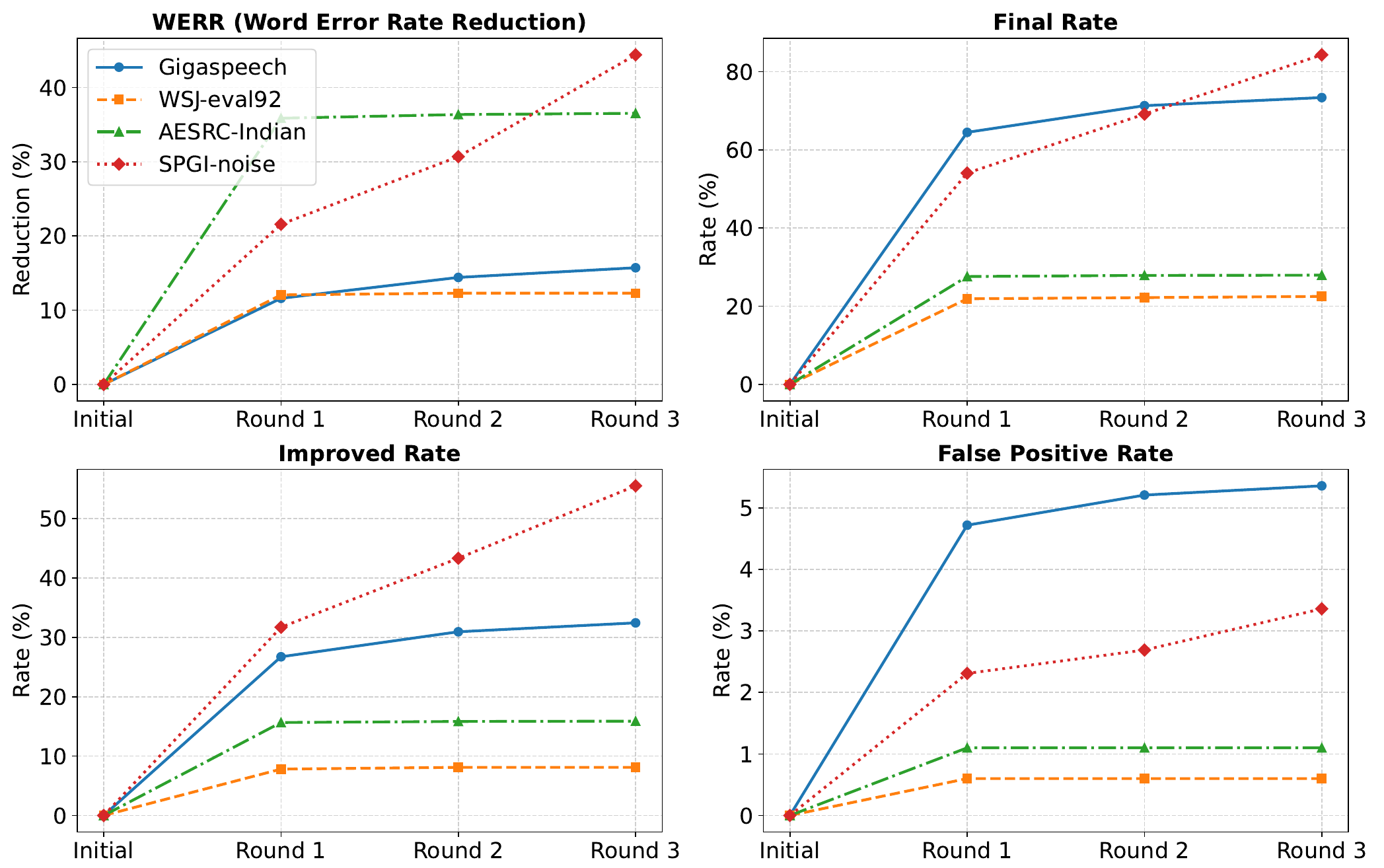}
    \caption{Theoretical performance limits of the $3$-round clarification loop under ideal feedback conditions (bypassing the user simulator's TTS/ASR).}
    \label{fig:k-round-details}
\end{figure*}

\subsection{Confusion Matrix for Distortion Event Detection}
\label{ap:event_detection}
Figure~\ref{fig:confusion_matrices} compares the confusion matrices for our distortion detection module at the acoustic \textit{frame level} (Figure~\ref{fig:confusion_frame}) versus the subword \textit{token level} (Figure~\ref{fig:confusion_token}).

\paragraph{Impact of Sampling Strategy.}
The shift from frame-level to token-level evaluation reveals how the detector performs at the critical time-steps that drive ASR output. While the global frame-level accuracy captures the detector's behavior across the entire audio duration (including silence and transitions), the token-level metrics isolate the instants of symbol generation. We observe that performance metrics shift based on this sampling; for instance, the recall for \textit{Signal Missing} drops from 91.1\% (frame) to 73.7\% (token), likely because true silence produces fewer tokens, and the few tokens emitted during missing segments often align with transient noise spikes that are harder to classify.

\paragraph{Conservative Detection Strategy.}
A consistent "conservative" bias is evident in both settings. The model frequently defaults to the \textit{Clean} class when distortion features are not overwhelming.
\begin{itemize}
    \item \textbf{Subtle Distortions (RIR, Packet Loss):} This bias is particularly pronounced for reverberation and packet loss. At the token level, 85.6\% of \textit{RIR} emission frames and 69.8\% of \textit{Packet Loss} emission frames are misclassified as \textit{Clean}. This indicates that at the exact moments of token emission, the acoustic features often retain enough clarity for the model to perceive them as "clean," despite the underlying distortion.
    \item \textbf{High Precision on Clean Speech:} Conversely, this conservative threshold ensures very high reliability for actual Clean speech, with 88.9\% accuracy at the token level, minimizing false alarms that could trigger unnecessary clarification turns.
\end{itemize}

\subsection{Analysis for K-round Clarification (theoretical performance)}

To assess the upper bound of our framework, we conducted a "theoretical performance" analysis shown in Figure~\ref{fig:k-round-details}. In this setting, the acoustic feedback loop is idealized: the User Simulator's response is fed directly into the Dialogue Manager as text, bypassing the TTS synthesis and ASR recognition steps. This isolates the error correction logic from potential recognition errors during the clarification turn itself.

\paragraph{Performance Potential (WERR \& Improved Rate).}
Eliminating acoustic noise from user feedback unlocks significant performance gains. As seen in the \textit{WERR} and \textit{Improved Rate} subplots, the theoretical reduction is substantially higher than the practical acoustic scenarios. Notably, \textbf{SPGI-noise} achieves a WERR of over 45\% (vs. $\sim$30\% in the practical setting), indicating that the dialogue manager correctly identifies and formulates strategies for complex noise-induced errors, provided the user's feedback is received clearly.

\paragraph{Convergence \& Complexity (Final Rate).}
The \textit{Final Rate} metric tracks the percentage of samples successfully resolved (clean transcript confirmed) at each round.
\begin{itemize}
    \item \textbf{Rapid Convergence:} \textbf{Gigaspeech} and \textbf{WSJ} exhibit a "fast saturation" behavior, where the majority of errors are resolved in Round 1 (Gigaspeech reaches $\sim$65\% Final Rate immediately). This suggests that errors in these domains are often simple misunderstandings resolvable with a single clarification.
    \item \textbf{Iterative Resolution:} In contrast, \textbf{SPGI-noise} shows a linear growth trend across all three rounds. This implies that environmental distortions create complex, multi-layered error spans that require iterative decomposition (e.g., first fixing the entity, then the context) rather than a "one-shot" fix.
\end{itemize}

\paragraph{Safety \& Stability (False Positive Rate).}
The \textit{False Positive Rate} (FPR) measures samples where the clarification process degraded the WER. Even in this theoretical setting, the FPR remains low ($\leq$ 1.2\%) for WSJ and AESRC. \textbf{Gigaspeech} shows a slightly higher FPR ($\sim$5.5\%), likely due to its spontaneous nature and ambiguous groundtruths, yet the trade-off remains highly favorable given the substantial WERR gains.

{
\subsection{Independent-Judge and Non-Cooperative-User Simulation}
\label{ap:sim_validity_details}
This section provides the full results behind Sec.~\ref{sec:sim_validity}.

\paragraph{Independent judge (Gemini-3.5-Flash).}
We re-scored all 21 frozen answer sets with \textit{Gemini-3.5-Flash} at temperature 0, using the byte-identical judge prompt of the GPT pass: 7 conditions $\times$ 3 runs $\times$ 200 answers (100 OpenHermes + 100 Alpaca) $=$ 4{,}200 samples, covering the four input conditions of Table~\ref{tab:dialogue_settings} (GT, ASR-Clean, ASR-Dist, Clarified) plus the three user-simulator profiles of Table~\ref{tab:ap_noncoop} (cooperative, mild, strong). Table~\ref{tab:ap_judge} reports the combined MaJ (mean over OpenHermes+Alpaca) under both judge families. Per-sample agreement is high: Pearson $r=0.895$, quadratic-weighted $\kappa=0.893$ (exact match 75.6\%), and $\kappa=0.923$ at the GT/Dist calibration endpoints. The GPT$-$Gemini delta is negative for \emph{every} condition family (Clarified $-2.37$, Clean $-2.57$, GT $-1.70$, Dist $-0.43$): GPT is uniformly the harsher judge on these GPT-generated answers, and the Clarified delta sits within the range of the other high-scoring families, so the headline recovery number is not selectively inflated by self-preference. Gap recovery is $0.78\pm0.15$ under GPT and $0.85\pm0.14$ under Gemini (mean $\pm$ 95\% CI, $n{=}3$).

\begin{table}[t]
    \centering
    \small
    \begin{tabular}{lcc}
        \toprule
        \textbf{Condition} & \textbf{GPT, N=3} & \textbf{Gemini, N=3} \\
        \midrule
        GT & 85.60{\scriptsize$\pm$0.90} & 87.30{\scriptsize$\pm$1.79} \\
        ASR-Clean & 84.93{\scriptsize$\pm$1.69} & 87.50{\scriptsize$\pm$0.43} \\
        ASR-Dist & 72.17{\scriptsize$\pm$2.08} & 72.60{\scriptsize$\pm$2.17} \\
        \textbf{Clarified} & \textbf{82.63{\scriptsize$\pm$2.45}} & \textbf{85.00{\scriptsize$\pm$2.40}} \\
        \bottomrule
    \end{tabular}
    \caption{Combined MaJ (mean over OpenHermes+Alpaca; mean $\pm$ 95\% CI, $n{=}3$) under the GPT judge and the independent Gemini-3.5-Flash judge, on identical frozen answer sets.}
    \label{tab:ap_judge}
\end{table}

\paragraph{Non-cooperative user simulators.}
The \textit{mild} simulator receives only an LLM-generated intent summary (never the reference transcript) and paraphrases imprecisely; the \textit{strong} simulator additionally has limited patience and may refuse follow-up questions or give short answers. Table~\ref{tab:ap_noncoop} shows that downstream MaJ is statistically indistinguishable across the three profiles. Two caveats apply: (i) these clarification turns are text-channel where the simulator's reply is fed directly to the Dialogue Manager without TTS/ASR,  which isolates user \emph{behavior} from the speech \emph{channel}; the channel is tested separately via accent-preserving TTS in Sec.~\ref{sec:sim_validity}; (ii) the intent summaries occasionally paraphrase content words, which elevates the mild/strong transcript WER upstream of the intended behavioral manipulation, yet the judged answer quality remains flat regardless.

\begin{table}[t]
    \centering
    \footnotesize
    \setlength{\tabcolsep}{2pt}
    \begin{tabular}{lccc}
        \toprule
        \textbf{User simulator} & \textbf{MaJ, N=3} & \textbf{$\Delta$ vs.\ coop.} & \textbf{$p$} \\
        \midrule
        Cooperative & 82.80{\tiny$\pm$0.86} & --- & --- \\
        Mildly uncoop. & 83.50{\tiny$\pm$3.52} & $+0.035$ {\tiny$[-0.043, +0.115]$} & 0.40 \\
        Strongly uncoop. & 82.60{\tiny$\pm$1.97} & $-0.010$ {\tiny$[-0.085, +0.068]$} & 0.82 \\
        \bottomrule
    \end{tabular}
    \caption{Downstream MaJ under cooperative and non-cooperative user simulators (mean $\pm$ 95\% CI, $n{=}3$). $\Delta$ is the paired per-sample bootstrap (10k resamples) of the 1--5 judge score pooled over 3 runs ($n{=}600$).}
    \label{tab:ap_noncoop}
\end{table}

\paragraph{Accent-preserving TTS.}
The main benchmark synthesizes all clarification replies with a fixed clean reference voice (Sec.~\ref{sec:hyper_params}), which removes part of the original acoustic difficulty. We therefore re-ran the full AESRC-Indian pipeline (N=3) with zero-shot voice cloning from the original accented speaker. A pilot ASR-proxy check on clarification-style sentences shows the clone retains part of the accent (WER: cloned 2.82\%, clean TTS 0.00\%, real accented speech ${\sim}6.08\%$). Final-round WER rises from $4.72\pm0.05$ under the clean fixed voice to $5.15\pm0.07$ under the cloned voice (Table~\ref{tab:wer_comparison}). Rank-matched per-utterance comparisons of the three run pairs give differences of $+0.61$, $+0.51$, and $+0.40$ WER (paired bootstraps, $p=0.0008$, $0.0018$, $0.0102$); pooling all per-utterance differences yields $+0.51$ WER, 95\% CI $[+0.31, +0.70]$. The clean-voice protocol thus overstates absolute recovery by ${\approx}0.5$ WER, while the direction and significance of the benefit are unchanged.

\subsection{Same-Backbone Non-Interactive Baseline}
\label{ap:nbest_baseline}
To isolate the value of interaction from the value of LLM post-correction, we compare our final-round results against a non-interactive, HyPoradise-style baseline built on the same Parakeet backbone: beam-4 N-best hypotheses corrected by a single \textit{gpt-5.2} call. Table~\ref{tab:ap_baseline} reports absolute WERs and per-utterance paired bootstraps (10k resamples) of the baseline against each of our three system runs. On clean, read speech (WSJ), the single-shot baseline is statistically indistinguishable from the interactive system ($+0.16\pm0.19$ WER, $p=0.18$--$0.69$ across runs): the residual errors there are largely language-model-recoverable, so one correction pass over the N-best list already captures them, and a user would gain little from clarification turns. The picture reverses under accent (AESRC, $+1.32\pm0.03$) and distortion (SPGI-noise, $+5.13\pm0.14$), where the difference is large and significant in every run ($p<10^{-4}$): there, the correct words are absent from the N-best list entirely, so no amount of single-shot correction can restore them; only asking the user can.

\begin{table}[t]
    \centering
    \scriptsize
    \setlength{\tabcolsep}{2pt}
    \begin{tabular}{lccccc}
        \toprule
        \textbf{Dataset} & \textbf{Init.$^{\S}$} & \textbf{N-best+LLM} & \textbf{Ours (N=3)} & \textbf{$\Delta$} & \textbf{$p$} \\
        \midrule
        WSJ-eval92 & 4.23 & 3.78 & 3.77{\tiny$\pm$0.17} & $+0.16${\tiny$\pm$0.19} & 0.18--0.69 \\
        AESRC-Indian & 6.08 & 5.84 & 4.72{\tiny$\pm$0.05} & $+1.32${\tiny$\pm$0.03} & $<10^{-4}$ \\
        SPGI-noise & 17.57 & 17.08 & 12.12{\tiny$\pm$0.22} & $+5.13${\tiny$\pm$0.14} & $<10^{-4}$ \\
        \bottomrule
        \multicolumn{6}{l}{\scriptsize{$^{\S}$Greedy one-best decoding of the backbone.}}\\
    \end{tabular}
    \caption{Same-backbone, non-interactive N-best+LLM post-correction baseline vs.\ our final-round WER. $\Delta$ = paired per-utterance mean difference (baseline $-$ ours; positive favors ours), mean $\pm$ 95\% CI over the three rank-matched comparisons; $p$ ranges over the per-run paired bootstraps.}
    \label{tab:ap_baseline}
\end{table}

\subsection{Attribution-Conditioned Recovery}
\label{ap:attribution}
To test whether misattribution breaks the recovery mechanism, we split per-utterance recovery (round-0 $\rightarrow$ round-3 $\Delta$WER, 3 runs) by whether the initial attribution matched the known injected cause (Table~\ref{tab:ap_attribution}). Attribution is judged against the injected cause, known by construction; $n$ counts error-bearing utterances per class and is identical across runs because the initial-turn decode and diagnosis are computed once and shared, and only the clarification turns vary. ``Typed'' denotes error spans tagged with the specific distortion-event type (e.g., \texttt{<noise>}); ``untyped'' denotes spans marked without an event-type label. Typed tagging both flags more error-bearing utterances (722 vs.\ 657 correctly attributed) and misattributes fewer (122 vs.\ 169).

Where misattribution is testable at scale (SPGI-noise), recovery degrades gracefully: wrongly attributed utterances still improve substantially, achieving roughly 60\% of the gain of correctly attributed utterances. The only net degradation is in the 4-utterance AESRC bucket, which is too small to generalize but is reported for completeness. The more consequential finding is the \emph{missed} class: utterances with recognition errors but no tag (67 for WSJ, 266 for AESRC, 322/304 for SPGI-noise per run) never enter clarification and pass through essentially unchanged ($\Delta$WER $\approx 0$). The mechanism's ceiling is therefore bounded by detection recall rather than by cause-family accuracy.

\begin{table}[t]
    \centering
    \scriptsize
    \setlength{\tabcolsep}{2pt}
    \begin{tabular}{lccc}
        \toprule
        \textbf{Dataset} & \textbf{Correct} & \textbf{Incorrect} & \textbf{Missed} \\
        \midrule
        AESRC-Indian & $-8.75${\tiny$\pm$0.22} & $+2.21${\tiny$\pm$4.20} {\tiny($n{=}4$)} & $-0.04$ \\
        SPGI-noise (untyped) & $-11.10${\tiny$\pm$0.27} & $-6.45${\tiny$\pm$0.95} {\tiny($n{=}169$)} & $0.00$ \\
        SPGI-noise (typed) & $-10.54${\tiny$\pm$0.39} & $-6.33${\tiny$\pm$1.84} {\tiny($n{=}122$)} & $0.00$ \\
        WSJ & $-4.17${\tiny$\pm$1.17} & --- & $0.00$ \\
        \bottomrule
    \end{tabular}
    \caption{Per-utterance $\Delta$WER (round-0 $\rightarrow$ round-3; mean $\pm$ 95\% CI over 3 runs) conditioned on whether the initial attribution matched the injected cause. ``Missed'' = error-bearing utterances that received no tag. WSJ has no misattributed utterances.}
    \label{tab:ap_attribution}
\end{table}
}

\subsection{Qualitative Evaluation of Clarification Naturalness}
\label{ap:naturalness}

To assess whether the system's clarification attempts feel natural and effective, we conducted a qualitative analysis using the Model-as-a-Judge (MaJ) framework on the Alpaca and OpenHermes evaluation sets.

\paragraph{Setup.}
We sampled all 141 clarification rounds generated during the 3-round interactive evaluation (Section~\ref{sec:sd_eval}). Each round was scored by GPT-5.2 along two binary dimensions: (1) \textbf{Naturalness}: whether the clarification question reads as a plausible, conversational follow-up rather than a mechanical prompt; and (2) \textbf{Effectiveness}: whether the clarification successfully elicited the information needed to correct the flagged error. We additionally report the distribution of overall MaJ quality scores (0--100) across all rounds.

\begin{table}[t]
    \centering
    \small
    \begin{tabular}{lcc}
        \toprule
        \textbf{Dimension} & \textbf{Pass Count} & \textbf{Pass Rate} \\
        \midrule
        Natural & 83 / 141 & 58.9\% \\
        Effective & 85 / 141 & 60.3\% \\
        \bottomrule
    \end{tabular}
    \caption{Binary naturalness and effectiveness evaluation of 141 clarification rounds on the Alpaca and OpenHermes evaluation sets, scored by the MaJ framework.}
    \label{tab:naturalness_binary}
\end{table}

\begin{table}[t]
    \centering
    \small
    \begin{tabular}{lcc}
        \toprule
        \textbf{Overall Assessment} & \textbf{Rounds} & \textbf{Percentage} \\
        \midrule
        90--100 (Excellent) & 18 & 12.8\% \\
        75--89 \phantom{0}(Good) & 67 & 47.5\% \\
        55--74 \phantom{0}(Mixed) & 40 & 28.4\% \\
        \phantom{0}0--54 \phantom{0}(Poor) & 16 & 11.3\% \\
        \bottomrule
    \end{tabular}
    \caption{Distribution of overall MaJ quality scores across 141 clarification rounds. Scores are binned into four tiers.}
    \label{tab:naturalness_distribution}
\end{table}

\paragraph{Results.}
Tables~\ref{tab:naturalness_binary} and~\ref{tab:naturalness_distribution} summarize the findings. Approximately 60\% of clarification attempts are judged both natural and effective, and the majority of rounds (60.3\%) receive an overall MaJ score of 75 or above (Good or Excellent). The remaining 40\% of rounds fall into the Mixed or Poor categories, often corresponding to cases where (i) the system's clarification question is overly verbose or formulaic, or (ii) the flagged error involves rare slang or proper nouns that resist targeted questioning.
\begin{figure}[tp]
    \centering
    \includegraphics[width=1.0\linewidth]{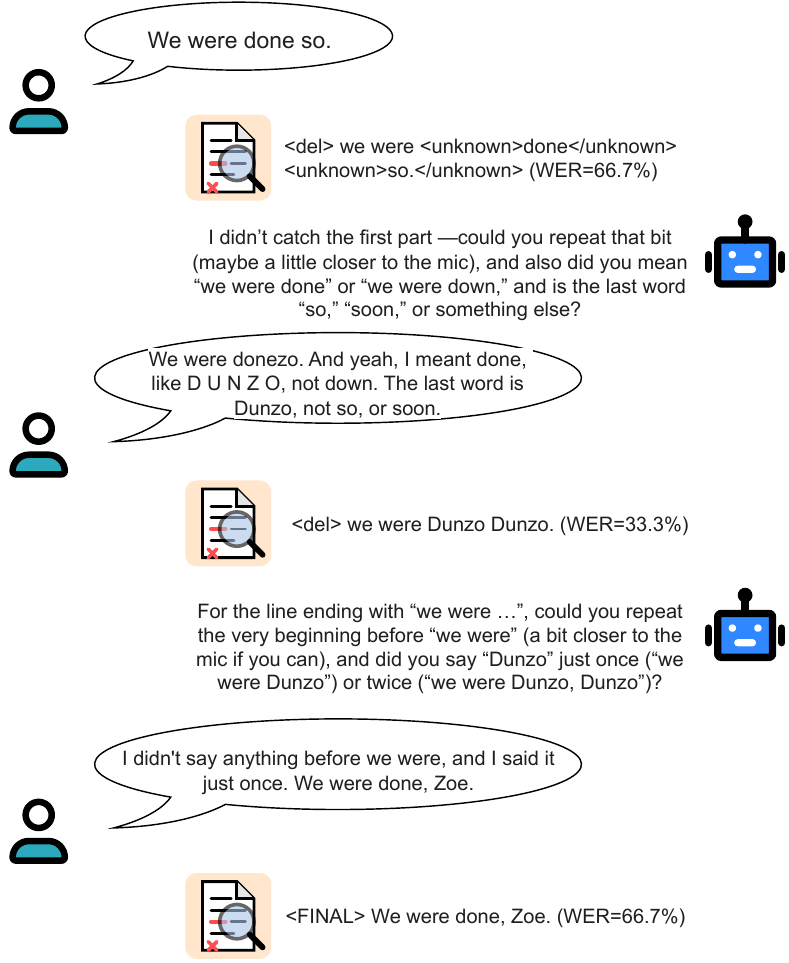}
    \caption{A failed case of iterative clarification. The groundtruth of the user input is: We were dunzo.}
    \label{fig:bad_sample}
\end{figure}

\begin{figure*}[tp]
    \centering
    \includegraphics[width=0.7\linewidth]{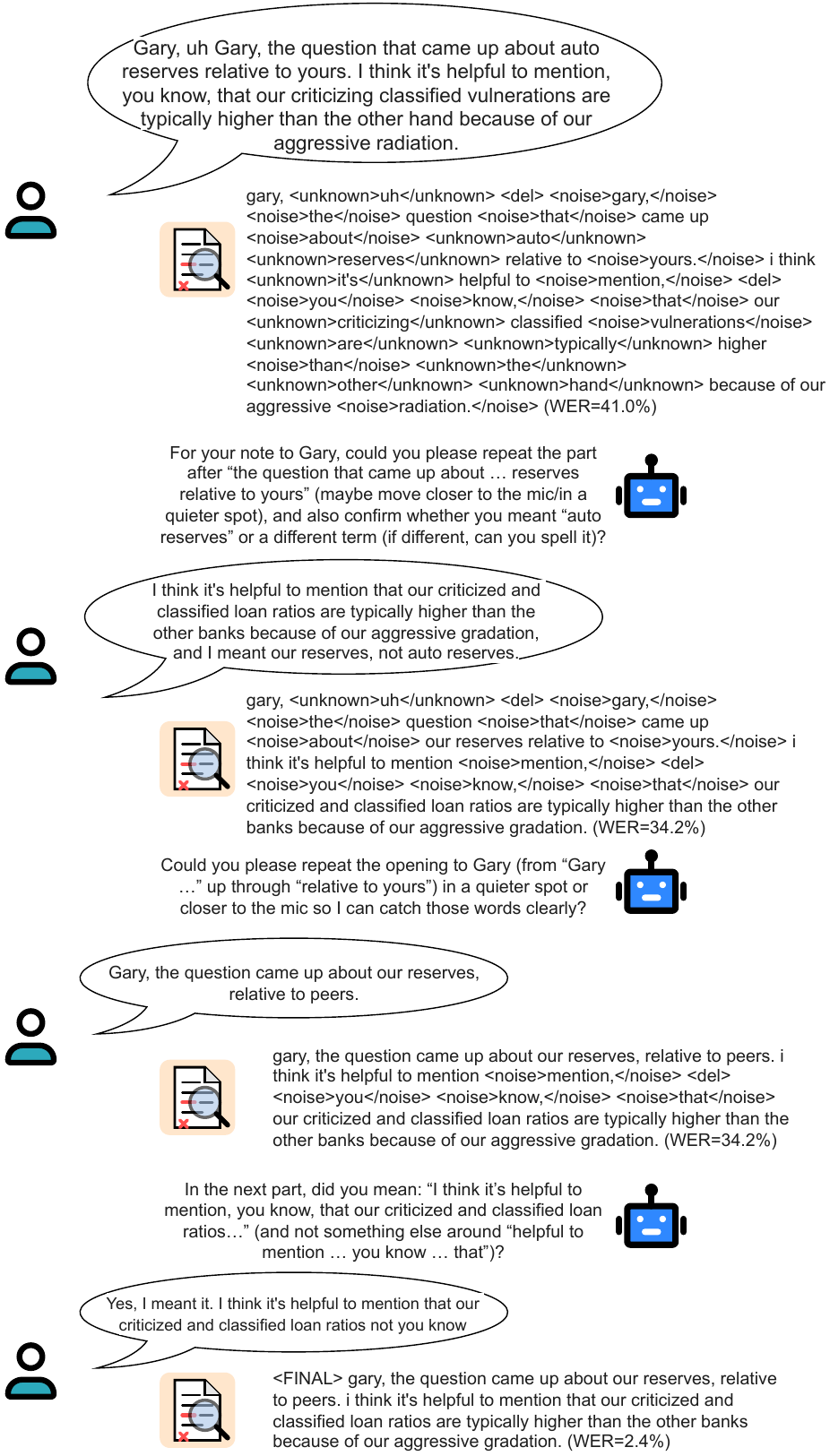}
    \caption{A good case of iterative clarification. The groundtruth of the user input is: Gary, the question came up about our reserves relative to peers. I think it is helpful to mention that our criticized and classified loan ratios are typically higher than the other banks because of our aggressive gradation.}
    \label{fig:good_sample}
\end{figure*}
\paragraph{Discussion.}
These results indicate that while cause-aware clarification substantially outperforms blind confidence filtering in resolving errors, the \textit{phrasing} of clarification questions remains a bottleneck. The 11.3\% Poor rate is concentrated in deletion-error scenarios, where the system must ask about content that was never transcribed: a fundamentally harder conversational task. We note that this evaluation is conducted with a simulated user; real users may exhibit more cooperative behavior (e.g., proactively offering context), potentially improving effectiveness. Nevertheless, these results highlight the need for future work on optimizing dialogue strategies, particularly to reduce verbosity and improve the pragmatic appropriateness of repair initiations.

\subsection{Case Study of K-round Clarification}

We present two examples of the system's performance in resolving speech-recognition errors. The first example (Figure~\ref{fig:bad_sample}) illustrates a failure case: despite the user explicitly spelling out the correction ("D-U-N-Z-O"), the system fails to integrate this constraint, resulting in a hallucinated entity ("Zoe") and a high final WER. The second example (Figure~\ref{fig:good_sample}) demonstrates a successful interaction: the system accurately identifies low-confidence segments within a complex financial statement (e.g., mistaking "gradation" for "radiation"). Through iterative, targeted questioning, the system correctly incorporates the user's feedback, progressively reducing the WER from 41.0\% to 2.4\%.

\end{document}